\documentclass[manuscript=article]{achemso}

\usepackage{siunitx}
\usepackage{graphicx}
\usepackage{subcaption}
\usepackage{amsmath}    
\usepackage{hyperref}
\usepackage{cleveref}   
\crefformat{equation}{#2eq.~#1#3}
\usepackage[version=4]{mhchem}

\newcommand{\smiles}[1]{\texttt{#1}}

\ExplSyntaxOn
\NewDocumentCommand{\tok}{m}
 {
    \seq_set_from_clist:Nn \l_tmpa_seq { #1 }
    \seq_indexed_map_inline:Nn \l_tmpa_seq {
        \int_compare:nF { ##1 = 1 } {\  }
        {
          \fboxsep=0.1em
          \fbox{\texttt{##2}}
        }
    }
 }
\ExplSyntaxOff

\newcommand{\labelphantom}[1]{%
  \parbox{0pt}{\phantomsubcaption\label{#1}}%
}

\let\oldcite\cite
\renewcommand{\cite}[1]{\unskip~\oldcite{#1}}

\newcommand{\datadrop}[1][doi:10.5281/zenodo.13761263]{\href{https://doi.org/10.5281/zenodo.13761263}{#1}}

\newcommand{\gh}[1]{\href{https://github.com/#1}{\includegraphics[height=11pt]{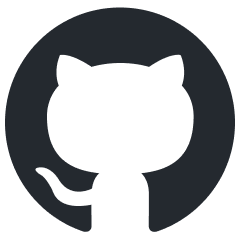}}}
\newcommand{\hf}[1]{\href{https://huggingface.co/#1}{\includegraphics[height=11pt]{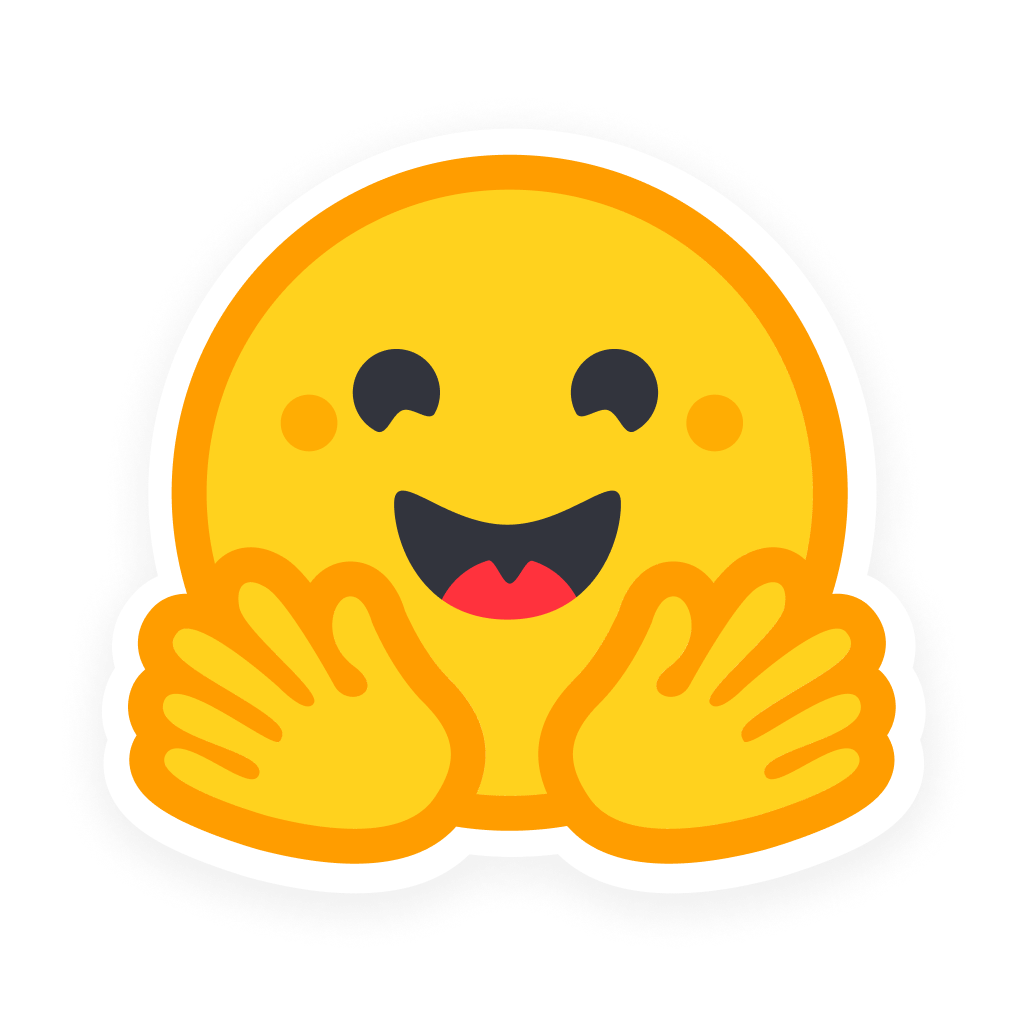}}}
\newcommand{\dd}[1]{\href{https://doi.org/10.5281/zenodo.13761263}{\includegraphics[height=11pt]{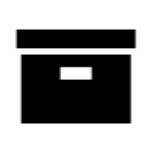}}}
\usepackage{acronym}

\acrodef{BPE}{Byte-Pair Encoding}
\acrodef{SPE}{SMILES Pair Encoding}
\acrodef{APE}{Atom Pair Encoding}
\acrodef{GPE}{Glyph Pair Encoding}
\acrodef{SMILES}{Simplified Molecular Input Line Entry System}
\acrodef{LLM}{Large Language Model}
\acrodef{NLP}{Natural Language Processing}
\acrodef{NMT}{Neural Machine Translation}
\acrodef{CE}{cross-entropy}
\acrodef{MAE}{mean absolute error}
\acrodef{RMSE}{root mean squared error}

\newcommand{\dropcap}[1]{#1}

\author{Alexius Wadell}
\affiliation[University of Michigan]
  {Department of Mechanical Engineering, University of Michigan, Ann Arbor, Michigan 48109, United States}
\alsoaffiliation[Carnegie Mellon University]
{Department of Mechanical Engineering, Carnegie Mellon University, Pittsburgh, Pennsylvania 15213, United States}
\author{Anoushka Bhutani}
\affiliation[University of Michigan]
  {Department of Mechanical Engineering, University of Michigan, Ann Arbor, Michigan 48109, United States}
\alsoaffiliation[Carnegie Mellon University]
{Department of Mechanical Engineering, Carnegie Mellon University, Pittsburgh, Pennsylvania 15213, United States}
\author{Venkatasubramanian Viswanathan}
\email{venkvis@umich.edu}
\affiliation[University of Michigan]
  {Department of Mechanical Engineering, University of Michigan, Ann Arbor, Michigan 48109, United States}
\alsoaffiliation[Carnegie Mellon University]
{Department of Mechanical Engineering, Carnegie Mellon University, Pittsburgh, Pennsylvania 15213, United States}

\title{Tokenization for Molecular Foundation Models}

\begin{document}

\dropcap{A}ccurate and fast prediction of chemical properties is essential to numerous industries from the design of next-generation energy storage devices\cite{AKM+AutoMatAutomatedMaterials2022,PPS+AcceleratingMaterialsDiscovery2022} to the discovery of pharmaceuticals \cite{SMV+MachineLearningSmall2023,ZXY+AccuratePredictionMolecular2022}.
Machine learning has emerged as a computationally tractable and accurate method to mitigate the high cost of experimentation or ab initio simulations\cite{MFC+HowMachineLearning2021}.
To this end, numerous machine learning techniques have been devised to accelerate material science computations, ranging from graph-neural networks\cite{ZNX+ImprovedEnvironmentalChemistry2023}, equivariant neural networks\cite{BMS+E3equivariantGraphNeural2022,ZBVPredictingElectrolyteConductivity2022}, and machine learning interatomic potentials\cite{PYB+AccurateSurfaceFiniteTemperature2024}.
With the success of the transformer architecture for \ac{NLP} \cite{VSP+AttentionAllYou2017,LOG+RoBERTaRobustlyOptimized2019,BMR+LanguageModelsAre2020}, recent efforts have sought to use the architecture for chemical property prediction\cite{SSB+LargeEncoderDecoderFamily2024,YUUDSELFormerMolecularRepresentation2023,XWBTransPolymerTransformerbasedLanguage2023,RBC+LargescaleChemicalLanguage2022,CGRChemBERTaLargeScaleSelfSupervised2020}, molecular design\cite{SSB+LargeEncoderDecoderFamily2024,BAVPMolGPTMolecularGeneration2022,BJJ+GenerativeModelsMolecular2022} and retrosynthetic analysis\cite{SLG+MolecularTransformerModel2019}, among others\cite{SGL+FoundTranslationPredicting2018,SVLRPredictionChemicalReaction2021,RCWReviewLargeLanguage2024}.

At their core, these efforts work by feeding molecules encoded as text into models designed for \ac{NLP}\cite{VSP+AttentionAllYou2017,LOG+RoBERTaRobustlyOptimized2019,BMR+LanguageModelsAre2020}.
The first step in this process is tokenization; this involves splitting the encoded molecules into short sequences of text or \emph{tokens}, which are then mapped to a finite set of integers (token IDs) using a \emph{vocabulary}\cite{MAS+WordsCharactersBrief2021}.
Early word-level tokenizers worked by splitting text into words and using a dictionary to look up their IDs\cite{MAS+WordsCharactersBrief2021}.
In turn, \acp{LLM} treat text as a probability distribution over tokens in their vocabulary, conditioned on the surrounding context\cite{BPX+LargeLanguageModels2007,VSP+AttentionAllYou2017,DCLTBERTPretrainingDeep2019,LOG+RoBERTaRobustlyOptimized2019}.
However, language models can only predict the probability of tokens within their vocabulary, causing problems when early models encountered novel words\cite{SHBNeuralMachineTranslation2016}.
In practice \emph{closed-vocabulary} tokenizers use a special unknown token \tok{[UNK]} to represent an unrecognized span of text\cite{SHBNeuralMachineTranslation2016,HuggingfaceTokenizersFast2024}.
Marking a span of text as unknown can be sufficient for some tasks, but is generally unsatisfactory\cite{SHBNeuralMachineTranslation2016,MAS+WordsCharactersBrief2021}.
Character-level tokenizers avoid the issue of unknown words by assigning a unique ID to each character, but higher inference costs have stymied their use\cite{MAS+WordsCharactersBrief2021,LBFishingMagikarpAutomatically2024}.
Instead, \emph{open-vocabulary} subword tokenizers, such as \ac{BPE} and others, have come to dominate existing \acp{LLM} \cite{MAS+WordsCharactersBrief2021,GagNewAlgorithmData1994,SHBNeuralMachineTranslation2016,SNJapaneseKoreanVoice2012,KudSubwordRegularizationImproving2018}.
By learning to break down words into multiple tokens, subword tokenizers seamlessly interpolate between a character and word-level tokenization\cite{SHBNeuralMachineTranslation2016,SNJapaneseKoreanVoice2012,KudSubwordRegularizationImproving2018}.

In contrast, molecular foundation models have converged on ``Atom-wise'' tokenization, a variant of word-level tokenization where encoded molecules are split into atom-level ``words'' and then mapped to token IDs using a fixed vocabulary\cite{SSB+LargeEncoderDecoderFamily2024,CGRChemBERTaLargeScaleSelfSupervised2020,RBC+LargescaleChemicalLanguage2022,SLG+MolecularTransformerModel2019,BAVPMolGPTMolecularGeneration2022,SKReactionT5LargescalePretrained2023,SPV+MappingSpaceChemical2021,SVLRPredictionChemicalReaction2021,SPV+MappingSpaceChemical2021,IDHBChemformerPretrainedTransformer2022,BADevalabMolgpt2023,YUUDSELFormerMolecularRepresentation2023}.
For molecules encoded using the \ac{SMILES}\cite{WeiSMILESChemicalLanguage1988} Schwaller et al.\cite{SGL+FoundTranslationPredicting2018} proposed the following widely used regular expression:

\begin{verbatim}
  (\[[^\]]+]|Br?|Cl?|N|O|S|P|F|I|b|c|n|o|s|p|\(|\)|\.
  |=|#|-|\+|\\\\\/|:|~|@|\?|>|\*|\$|\%[0-9]{2}|[0-9])
\end{verbatim}

Critically, the leading pattern \texttt{\textbackslash[[\^{}\textbackslash]]+\textbackslash]} treats all ``bracketed atoms'' as a single irreducible token.
Bracketed atoms represent any atom outside the organic subset \tok{B,C,N,O,S,F,Cl,Br,I} or atoms with an explicit nuclear, geometric, or electronic aspect (\cref{fig:smiles_syntax}).
According to the OpenSMILES specification, these aspects are encoded into bracketed atoms as \texttt{bracket\_atom ::= `[' isotope? symbol chiral? hcount? charge? class? `]'}\cite{CraOpenSMILES2016}.
By treating each permutation as unique, Atom-wise tokenizers would require an extremely large vocabulary, in excess of 28 trillion tokens, for full coverage of the OpenSMILES specification.
However, current Atom-wise tokenizers have fewer than three thousand tokens, resulting in significant gaps in their coverage~(\cref{fig:oov_tokens}).

\begin{figure*}
  \centering
  \labelphantom{fig:smiles_syntax}
  \labelphantom{fig:example_tokenization}
  \labelphantom{fig:oov_tokens}
  \includegraphics[width=\linewidth]{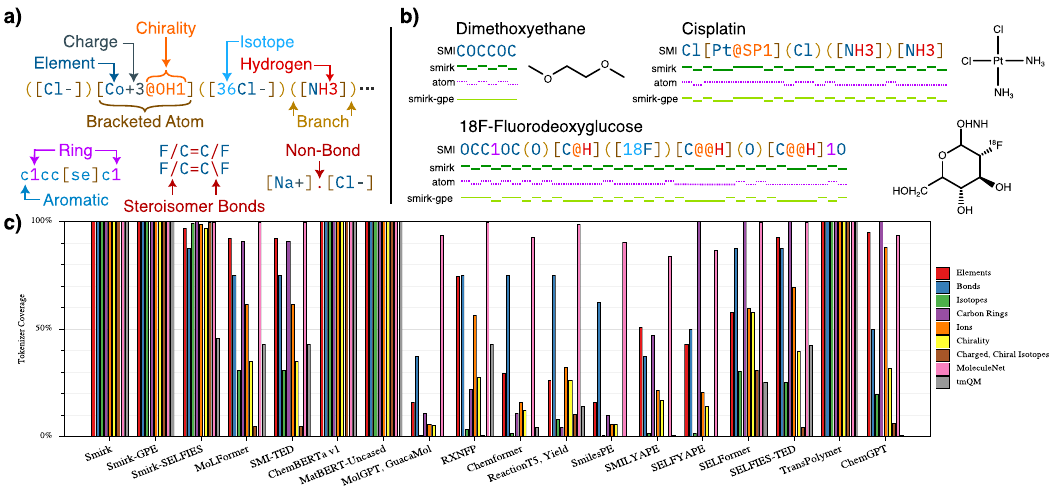}
  \caption{
    \subref*{fig:smiles_syntax})~In SMILES, bracketed atoms are information-rich, encoding isotopes, chiral centers, charge, species, and hydrogen bonds.
    Atom-wise tokenizers emit a single token per bracketed atom, requiring an extremely large vocabulary size to cover each possible combination of features.
    Smirk avoids this by fully decomposing the bracketed atoms.
    Smirk-GPE goes a step further by using a variant of BPE to compress the tokenized sequence.
    \subref*{fig:example_tokenization})~Example tokenization represented by a sequence of dashes, where each dash represents an emitted token.
    Atom-wise tokenizers emit fewer tokens than Smirk for bracketed atoms but run the risk of out-of-vocabulary tokens.
    Smirk-GPE balances the two by merging common snippets into a single token and falling back to a verbose encoding as required.
    \subref*{fig:oov_tokens})~Existing Atom-wise tokenizers lack complete coverage of the OpenSMILES specification, with only the open-vocabulary tokenizers used by ChemBERTa and TransPolymer achieving full coverage.
    Transcoding errors, when converting from SMILES\cite{WeiSMILESChemicalLanguage1988} to SELFIES\cite{KHN+SelfreferencingEmbeddedStrings2020}, counted against coverage.
  }
\end{figure*}

While the vast majority of molecular foundation models use Atom-wise tokenization, there are some notable variations between models.
IBM's MoLFormer and SMI-TED use nearly identical vocabularies constructed from their pretraining datasets\cite{RBC+LargescaleChemicalLanguage2022,SSB+LargeEncoderDecoderFamily2024}.
Conversely, MolGPT used separate vocabularies for the MOSES and GuacaMol benchmarks\cite{BAVPMolGPTMolecularGeneration2022}.
\ac{SPE} and \ac{APE} are two novel chemistry-specific tokenization schemes fusing Atom-wise and \ac{BPE} tokenization\cite{LFSMILESPairEncoding2021,LPP+ComparingSMILESSELFIES2024}.
Starting from tokens produced by the Atom-wise regular expression, these tokenizers learn a set of merge rules to combine adjacent tokens, akin to \ac{BPE}.
However, by starting from the non-atomic Atom-wise tokens, both methods remain closed-vocabulary models unlike \ac{BPE}.
ChemBERTa\cite{CGRChemBERTaLargeScaleSelfSupervised2020}, TransPolymer\cite{XWBTransPolymerTransformerbasedLanguage2023}, ReactionT5\cite{SKReactionT5LargescalePretrained2023} and SELFormer\cite{YUUDSELFormerMolecularRepresentation2023} all use open-vocabulary tokenizers with varying degrees of customization for their task.
TransPolymer leveraged RoBERTa's pretrained BPE tokenizer directly, while the other three trained their own variants.
Unfortunately, only the ChemBERTa and TransPolymer vocabularies contain a complete alphabet, a prerequisite for open-vocabulary modeling.
Both SELFormer and ReactionT5 omit tokens for \tok{U} or \tok{u}, preventing the representation of  copper (\smiles{Cu}), ruthenium (\smiles{Ru}), gold (\smiles{Au}), europium (\smiles{Eu}), lutetium (\smiles{Lu}), uranium (\smiles{U),} and plutonium (\smiles{Pu}).

\subsection{Smirk}
To enable the complete coverage of the OpenSMILES\cite{CraOpenSMILES2016} specification, we propose fully decomposing the bracketed atoms into their consistent glyphs using a two-stage tokenization scheme.
First decomposing a \ac{SMILES} encoding into atoms (\(\smiles{OC[C@@H][OH]} \to \tok{O,C,[C@@H],[OH]}\)) and then into its constituent glyphs (\tok{O,C,[,C,@@,H,],[,O,H,]});
regular expressions for both stages are provided in the supporting information.
In essence, Smirk is a character-level tokenizer operating over the glyphs defined by OpenSMILES instead of the Unicode Consortium.
The two-step process is necessary to distinguish between, for example, \smiles{Sc} representing a sulfur-carbon bond and \smiles{[Sc]} for scandium -- a seemingly esoteric ambiguity that occurs over half a million times within PubChem's compound dataset, as detailed in our supporting information for details.

The resulting vocabulary consists of 165 tokens requires no training and by construction can faithfully tokenize any OpenSMILES encoded molecule (\cref{fig:oov_tokens}).
We have implemented the proposed tokenization scheme in Rust using HuggingFace's Tokenizers\cite{HuggingfaceTokenizersFast2024} library and have made the code and prebuilt-wheels openly available at \url{https://github.com/BattModels/Smirk} and on PyPI.
We have also implemented an equivalent tokenization scheme (\emph{Smirk-SELFIES}) for SELFIES\cite{KHN+SelfreferencingEmbeddedStrings2020} encoded molecules.

\subsection{Glyph Pair Encoding}
The above scheme is not without its drawbacks.
The average tokenized sequence length, or fertility, is higher for Smirk than current Atom-wise tokenizers; using at least two more tokens for any bracketed atom (\tok{[,Au,]} versus \tok{[Au]}).
Unfortunately, a longer sequence length increases the computational cost of attention, which grows quadratically with sequence length\cite{CFB+RevisitingCharacterBasedNeural2018,LFSMILESPairEncoding2021,LPP+ComparingSMILESSELFIES2024}.

To mitigate this regression, we implemented \emph{Smirk-GPE}, which further compresses Smirk's tokenization using a variant of \ac{BPE} that operates on chemically-meaningful glyphs rather than bytes or characters\cite{GagNewAlgorithmData1994,SHBNeuralMachineTranslation2016,HuggingfaceTokenizersFast2024,OpeOpenaiTiktoken2024}.
As with \ac{BPE}, merges between adjacent tokens are replaced with a single meta-token using merge rules learned from a training corpus\cite{GagNewAlgorithmData1994,SHBNeuralMachineTranslation2016}.
Unlike \ac{BPE}, these merge rules operate on token IDs rather than pairs of strings, ensuring that \(\mbox{merge}(\mbox{\tok{S}}, \mbox{\tok{c}})\) -- representing a sulfur–carbon bond -- remains distinct from the atomic symbol token \tok{Sc} for Scandium.
\ac{SPE} and \ac{APE} tokenizers avoided the need to handle this ambiguity by starting from Atom-wise tokens, however \ac{BPE} tokenizers are susceptible.
For instance, ChemBERTa uses the same \tok{Sc} token to tokenize both \smiles{[Sc]} (Scandium) and \smiles{Cn1nccc1Sc1ccccc1}\cite{cid101490041};
a similar reuse occurs for Copernicium with \smiles{[Cn]}.
This can lead to distinct chemical entities being conflated during downstream analysis.
Similar issues arise when comparing the \smiles{OH} in \smiles{[OH]} and \smiles{[C@OH1]}.
The former \smiles{OH} is an oxygen-hydrogen bond, while the latter indicates a carbon octahedral chiral center.
The downstream impact of these ambiguities remains unclear, especially in light of the empirical performance of ChemBERTa and \acp{LLM} at large.

\section{Evaluating Tokenizers for Chemistry}
Tokenization plays a foundational role in language modeling, yet evaluating its impact remains a nascent area of research\cite{BVD+EvaluatingSubwordTokenization2024,GCE+UnpackingTokenizationEvaluating2024,AFT+TokenizerChoiceLLM2024,MAS+WordsCharactersBrief2021}.
Evaluating tokenizers predominately relies on expensive extrinsic metrics (such as downstream benchmarks) of the complete (pre-trained and finetuned) language model\cite{GCE+UnpackingTokenizationEvaluating2024,RPV+HowGoodYour2021,AFT+TokenizerChoiceLLM2024}.
Intrinsic tokenizer metrics, which evaluate the tokenizer in isolation are critical to mitigating this cost, as evidenced by Google's DeepMind researchers lamenting \emph{their limited compute budget} to conduct this line of research\cite{GCE+UnpackingTokenizationEvaluating2024}.

Tokenizer metrics may be nascent, but the deleterious effects of poor tokenization are well documented\cite{AFT+TokenizerChoiceLLM2024,RPV+HowGoodYour2021,SSTokenizationCountsImpact2024}.
Poor tokenizer design has been linked to impaired multilingual performance\cite{AFT+TokenizerChoiceLLM2024,RPV+HowGoodYour2021} leading to both higher-costs and reduced quality for non-Latin script languages\cite{AKG+AllLanguagesCost2023}.
In the scientific domain, Lindsey et al.\ found tokenizer choice had a significant impact on the performance of both attention and state-space based genomic models\cite{LPH+ComparisonTokenizationImpact2024}.
Chithrananda et al.\ noted that Atom-wise tokenization provided a slight advantage over \ac{BPE} on ToxCast, but ultimately selected a \ac{BPE} tokenizer for ChemBERTa\cite{CGRChemBERTaLargeScaleSelfSupervised2020}.

\begin{figure}
  \centering
  \includegraphics[width=\linewidth]{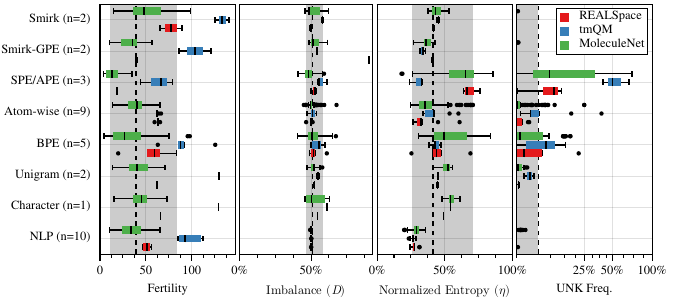}
  \caption{\label{fig:intrinsic}
    Multiple intrinsic metrics have been developed to assess tokenizer quality in isolation including fertility\cite{JudExploringBERTVocabulary2019,GMFindingOptimalVocabulary2020}, imbalance \cite{GMFindingOptimalVocabulary2020}, normalized entropy and frequency of the unknown token.
    We have tabulated all four metrics for 34 tokenizers on REALSpace\cite{EnaREALSpace2024}, MoleculeNet\cite{WRF+MoleculeNetBenchmarkMolecular2018} and tmQM\cite{BSTmQMDatasetQuantum2020}.
    For clarity, we have categorized chemistry-specific tokenizers by tokenization scheme and grouped all \ac{NLP} tokenizers regardless of method;
    the number of tokenizers for each class is indicated in the parenthesis next to the class name ($n = \textrm{num tokenizers}$).
    Metric means and 90\% quantiles are indicated by vertical lines and shaded regions, respectively.
    Tokenizer specific metrics are provided in our supporting information.
  }
\end{figure}

One intrinsic metric of tokenizer quality is fertility, the mean tokenized sequence length averaged over words\cite{JudExploringBERTVocabulary2019,GMFindingOptimalVocabulary2020}, corpus\cite{LPH+ComparisonTokenizationImpact2024} or molecules\cite{LPP+ComparingSMILESSELFIES2024,LFSMILESPairEncoding2021}.
At a minimum, fertility captures the quadratic scaling of compute costs with sequence length due to dot-product attention\cite{AFT+TokenizerChoiceLLM2024}.
With both \ac{SPE} and \ac{APE} tokenizers highlighting their lower fertility and resulting cost reduction relative to Atom-wise tokenization\cite{LPP+ComparingSMILESSELFIES2024,LFSMILESPairEncoding2021}.
That is, all things being equal, doubling the fertility of a tokenizer would approximately quadruple a model's training and inference cost.
Additionally, prior works have found a correlation between higher fertility and worse downstream model performance\cite{GCE+UnpackingTokenizationEvaluating2024,RPV+HowGoodYour2021,GMFindingOptimalVocabulary2020}.
In particular, Goldman et al.\ observed that more compressive tokenizers tend to increase the negative-log-likelihood of the text, effectively aligning with the training objective of \acp{LLM}\cite{GCE+UnpackingTokenizationEvaluating2024}.
On this basis, Smirk tokenization is a regression in quality relative to existing tokenizer (\cref{fig:intrinsic}), particularly on the tmQM\cite{BSTmQMDatasetQuantum2020} dataset due to its abundance of bracketed atoms.
Building on Goldman et al.'s analysis, we propose using normalized entropy (\cref{eq:normalized_entropy}) to measure the compression provided by a tokenizer.
Normalized entropy \(\eta\) evaluates how close a tokenizer comes to the information-theoretic ideal, where all tokens are equally probable\cite{ShaMathematicalTheoryCommunication1948,AllIndicesQualitativeVariation1967} and is defined as:

\begin{equation}\label{eq:normalized_entropy}
  \eta = \frac{-1}{\log |V|} \sum_{x \in V} p(x) \log p(x)
\end{equation}

\noindent
where \(V\) is the tokenizer's vocabulary and \(p(x)\) gives the probability for each token \(x \in V\) as observed in the corpus.
As with fertility, normalized entropy, does not capture the effect of an open versus closed vocabulary tokenizer, instead evaluating how a tokenizer has allocated its vocabulary.
Following Goldman et al.'s\ analysis, tokenizer performance should increase with normalized entropy.
Notably, \ac{NLP} \ac{BPE} tokenizers score poorly here with \(\eta \approx 25\%\) while \ac{SPE}/\ac{APE} tokenizers score highly (\cref{fig:intrinsic}).
Gowda and May proposed a related measure \(D = \frac{1}{2}\sum_{x\in V} \left| p(x) - |V|^{-1} \right|\) to measure the token imbalance present within a dataset\cite{GMFindingOptimalVocabulary2020}.
All tokenization schemes score similarly with \(D \approx 50\%\);
except for Smirk-GPE on tmQM (\cref{fig:intrinsic}) indicating the merges Smirk-GPE learned from REALSpace did not generalize to this dataset.
Both measures (\(D\) and \(\eta\)) measure the distance between observed token probabilities \(p(x)\) and a uniform distribution over the vocabulary.
Overall, all tokenizers score similarly on all three intrinsic metrics with nearly all scores sitting within the 90\% quantile (\cref{fig:intrinsic}).

Unfortunately, unlike the \ac{NLP} tokenizers for which existing metrics were developed\cite{JudExploringBERTVocabulary2019,GMFindingOptimalVocabulary2020} many chemistry-specific tokenizers are not open-vocabulary.
As such, fertility, normalized entropy (\(\eta\)) and token imbalance (\(D\)) all miss vocabulary specific issues pertinent to existing chemistry-specific tokenizers.
\ac{APE} and \ac{SPE} score extremely well on all three metrics, but their low coverage (\cref{fig:oov_tokens}) results in the unknown token being \(18.9\%\) of their emitted tokens on MoleculeNet and \(\approx 50\%\) on tmQM.
In fact, all existing chemistry specific tokenizer (SPE/APE, Atom-wise, BPE and Unigram) emit the unknown token with a non-negligible frequency (\cref{fig:intrinsic}).
Conversely, existing \ac{NLP} tokenizers (i.e. GPT-4o, LLama, Gemma, etc.), Smirk and Smirk-GPE do not, while scoring similarly to chemistry-specific tokenizers on all other intrinsic metrics.

\subsection{Low-Cost Proxy Language Model}

\begin{figure}
  \centering
  \includegraphics[width=\linewidth]{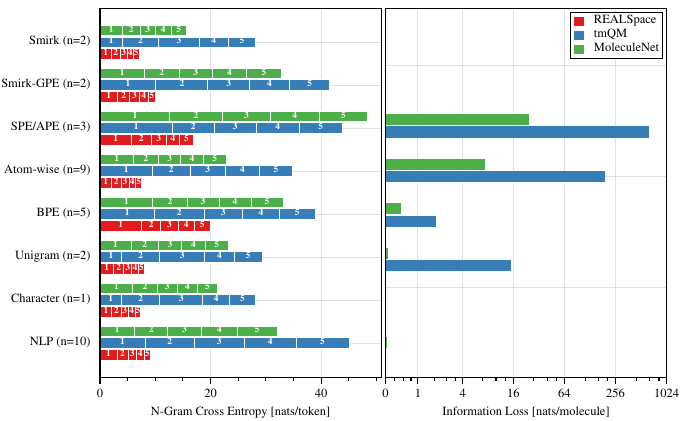}
  \caption{
    \label{fig:ngram_metrics}
    N-Gram cross-entropy and information loss as evaluated on the validation splits of our three datasets.
    As expected, the cross-entropy loss decreases with increasing n-gram order, since the added context reduces the uncertainty of the next token\cite{ShaPredictionEntropyPrinted1993}.
    Overall, we find Smirk maintains a lower cross-entropy loss when moving from pretraining (REALSpace) to downstream (MoleculeNet and tmQM) tasks;
    this suggests the learned language model was more applicable to the downstream tasks.
    This is also true for existing \emph{open-vocabulary} \ac{NLP} tokenizers which score a cross-entropy on par with existing chemistry-specific tokenizers.
    Conversely, existing chemistry-specific tokenizers lose a non-negligible amount of information to the unknown token regardless of tokenization scheme (\ac{SPE}, \ac{APE}, \ac{BPE}, Atom-wise or Unigram).
    Tokenizer-specific metrics are tabulated in the supporting information.
  }
\end{figure}

Intrinsic tokenizer metrics are well-suited for evaluating vocabulary sizing and mono/multilingual performance\cite{RPV+HowGoodYour2021,WSF+BLOOM176BParameterOpenAccess2023,GMFindingOptimalVocabulary2020}.
However, as discussed above, most existing chemistry-specific tokenizers are predominantly closed-vocabulary, which raises downstream performance concerns that are highly dependent on vocabulary coverage.
For instance, omitting the \tok{C} token would almost certainly be catastrophic for an organic chemistry foundation model, yet even OpenSMILES lacks support for Oganesson\cite{CraOpenSMILES2016}.
While the limited coverage of chemistry-specific tokenizers may be deleterious, the impact on model quality could be negligible if the surrounding context provides sufficient information to infer the content of the obscured text.
Probing these questions using a transformer model rapidly become computationally intractable, while confounding variables (i.e. hyperparameter selection, model architecture, etc.) reduce the power of any analysis.

To address this, we propose using the original ``large language model''\cite{BPX+LargeLanguageModels2007}, the n-gram, as a low-cost proxy for transformer-based models.
Similar to transformer-based models, n-grams estimate the likelihood of a token \(x_i\), given the preceding \(n-1\) tokens \(x_{i-n+1}, \dots, x_{i-1}\):

\begin{equation} \label{eq:ngram_next_token}
P_n(x_i| x_{i-n+1}, \dots, x_{i-1}) = \frac{C(x_{i-n+1}, \dots, x_{i}) + 1}{ C(x_{i-n+1}, \dots, x_{i-1}) + |V|}
\end{equation}

\noindent
where \(P_n\) is the likelihood of \(x_i\) being the \(n^{th}\) token, \(C\) gives the count of the n-gram in the training corpus, and \(|V|\) is the vocabulary size.
After ``pretraining'' n-grams on 1.6B \ac{SMILES} from Enamine REAL Space\cite{EnaREALSpace2024} we evaluated their cross-entropy loss on the pretraining and MoleculeNet validation splits (\cref{fig:ngram_metrics}).
Cross-entropy loss measures the distance between the distribution of the model and the data and is the primary metric used to train language models\cite{VSP+AttentionAllYou2017,DCLTBERTPretrainingDeep2019}.
As such, n-grams precisely proxy the training and evaluation of a decoder-only transformer model for only a fraction of the compute costs.

\subsection{Extrinsic Metrics}

\begin{figure*}
  \centering
  \labelphantom{fig:effect_size}
  \labelphantom{fig:ngram_prog}
  \includegraphics[width=\linewidth]{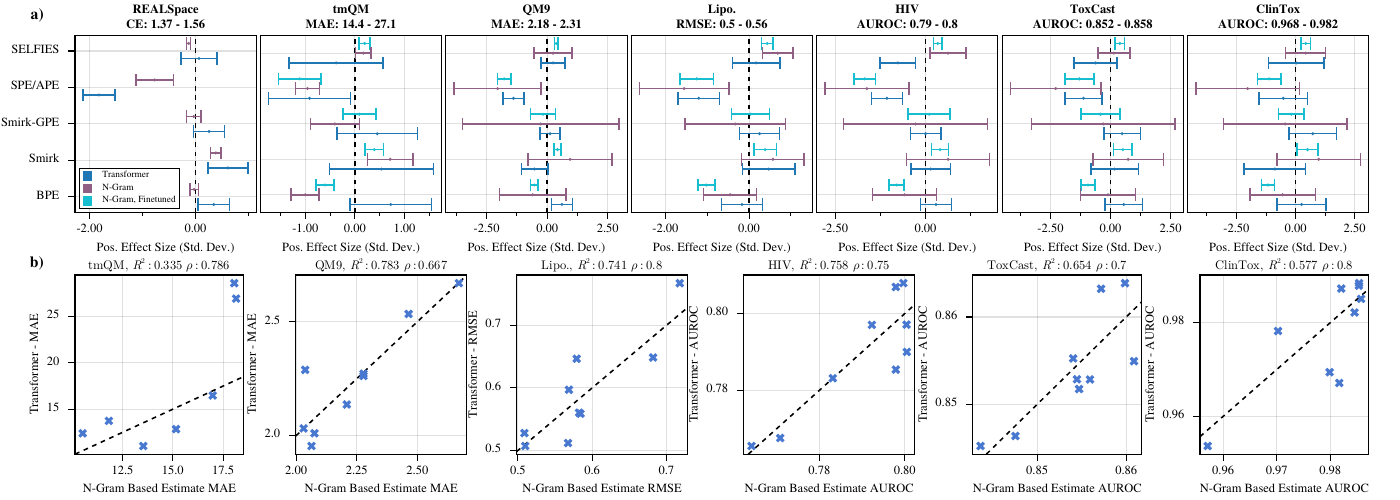}
  \caption{
    \label{fig:prognostic}
    \subref*{fig:effect_size})~Standardized effect sizes for different tokenization schemes and molecular encodings, estimated using fixed-effects models relative to an Atom-wise and \ac{SMILES} baseline.
    The 90\% quantiles for baseline model quality are shown above each subfigure.
    For clarity, the sign of \ac{CE}, \ac{MAE} and \ac{RMSE} effects have been flipped so that improvements are consistently positive.
    \subref*{fig:ngram_prog})~N-gram statistics -- cross-entropy and information loss -- are linearly predictive of downstream molecular foundation model performance.
    Spearman's rank correlation coefficient (\(\rho\)) consistently indicates that n-gram metrics capture the majority of the rank correlation, demonstrating their utility for estimating the relative performance of different tokenization schemes.
  }
\end{figure*}

To validate n-grams as a low-cost proxy for transformer based language models, we pretrained 18 encoder-only RoBERTA models spanning 11 tokenizers and three molecular encoding.
Critically, these models were trained from scratch using tokenizers from existing molecular foundation models, thereby isolating the effect of the tokenizer.
Detailed information on the training protocol and model hyperparameters can be found in \cref{sec:training_protocol} of our Material and Methods.
We finetuned each model on six regression and seven classification tasks from MoleculeNet\cite{WRF+MoleculeNetBenchmarkMolecular2018} and tmQM\cite{BSTmQMDatasetQuantum2020}.
We used linear fixed-effects models to evaluate the impact of tokenizer class and molecular encoding, relative to an Atom-wise tokenization and \ac{SMILES} encoding baseline (\cref{fig:effect_size}).
Overall, we found the use of \ac{APE} or \ac{SPE} tokenizers to have negative impact on both pretraining and downstream performance relative to our baseline.
Smirk has a positive effect on pretraining quality and tmQM downstream performance, but performs similarly to Atom-wise tokenization on MoleculeNet tasks.
We consistently found the choice of molecular encoding to be negligible.

Overall, we find transformer and n-gram effect sizes to be directionally consistent with each other, supporting our use of n-grams as lower-cost proxy model.
Assessing downstream performance using the pretrained or finetuned n-gram models showed similar effect sizes (\cref{fig:effect_size});
the finetuned variant reduced the effect size variance, but generally did not shift the expectation.
Extrinsic metrics should remain the gold standard for evaluating the impact of tokenization on the quality of the resulting large language model.
However, our findings suggest that n-gram models reliably offer a lower-cost alternative to supplement these costly metrics.

\subsection{Information Loss from Unknown Tokens}
We hypothesize that complete coverage of the OpenSMILES specification is desirable for a molecular language model.
However, current models have achieved competitive performance despite lacking full support for the specification (\cref{fig:oov_tokens,fig:intrinsic}).
This is partially a reflection of the current benchmarks used to evaluate molecular foundation models.
For example, our corpus of 2.1B molecules lacks any molecules containing quadruple bonds.
This paradox could also be explained by the fact that the information lost to unknown tokens may simply be negligible.
For example, a tokenizer lacking a token for oxygen (\tok{O}) would tokenize \smiles{C1CCOC1} as \tok{C, 1, C, C, [UNK], C, 1}.
If oxygen were the only non-carbon atom in heterocyclic compounds, no information would be lost by replacing the \tok{O} with the \tok{[UNK]} token.

To quantify the information lost to unknown tokens, we computed the KL-divergence between the distribution with (\(B'_n\)) and without (\(B_n\)) unknown tokens.
To proxy the bidirectional context of encoder-only transformer models, we used a character n-gram model to compute \(B_n\) as the probability of a token \(x_i\) from the joint probability of the preceding and succeeding \(n-1\) tokens:

\begin{multline} \label{eq:ngram_mlm}
B_n(x_i | x_{i-n+1}, \dots x_{i-1}, x_{i+1}, \dots, x_{i+n-1}) \propto \\
  \frac{C(x_{i-n+1}, \dots, x_{i}) + 1}{ C(x_{i-n+1}, \dots, x_{i-1}) + |V|}
  \times
  \frac{C(x_{i}, \dots, x_{i+n-1}) + 1}{ C(x_{i+1}, \dots, x_{i+n-1}) + |V|}
\end{multline}

Next, we marginalize \(B_n\) over the tokens that would be marked as unknown by the tokenizer under evaluation to get \(B'_n\).
The KL-divergence between \(B_n\) and \(B'_n\) is then the information lost to the unknown tokens: \(D_{KL}(B_n || B'_n) = \sum B \cdot \left( \log B_n - \log B'_n \right)\).
A detailed derivation of \(B_n\) and \(B'_n\) and their computation is presented in the supporting information.

Using a reference n-gram model and tokenizer is necessary to provide a baseline for comparison, as the tokenizer under evaluation does not have any other way to represent the unknown span of text.
As a data-driven method, the reference n-gram model is limited by the distribution of the training corpus.
We used a character-level 5-gram model as our reference tokenizer, as we could precisely marginalize over unknown tokens and its performance was on par with other models (\cref{fig:ngram_metrics}).

We found that information loss was minimal for tokenizers with robust dataset coverage (\cref{fig:ngram_metrics}).
In contrast, tokenizers with limited coverage exhibited substantial losses, particularly on datasets such as tmQM.
We selected tmQM~\cite{BSTmQMDatasetQuantum2020} for its broader range of elements and stereochemical centers (e.g., \smiles{[Co@OH5]}), both of which are encoded using bracketed atoms.
By comparison, MoleculeNet and REALSpace primarily consist of small organic molecules composed of a limited set of elements, with minimal stereochemical complexity or other bracketed-atom features (\cref{fig:smiles_syntax}).
For example, MoLFormer~\cite{RBC+LargescaleChemicalLanguage2022} incurs only \(0.1\) nats/molecule of information loss on MoleculeNet, but suffers a loss of \(40.3\) nats/molecule on tmQM, as unknown tokens increasingly obscure critical information.
In contrast, Smirk, Smirk-GPE, and other open-vocabulary tokenizers mitigate this degradation across both datasets, yielding measurable gains on challenging downstream tasks (\cref{fig:effect_size}).
These findings underscore the practical advantages of robust coverage, particularly for handling chemically rich syntax as found in tmQM.

\section{Discussion}
The two new tokenizers, Smirk and Smirk-GPE, introduced in this work can represent the entirety of the OpenSMILES\cite{CraOpenSMILES2016} specification.
In contrast to prior works, Smirk splits bracketed atoms into their constituent glyphs, enabling full coverage with a limited vocabulary (\cref{fig:oov_tokens})
We also implemented Smirk-GPE, a \ac{BPE}-like tokenizer, to further compress the Smirk tokenization -- akin to earlier \ac{APE} and \ac{SPE} tokenizers\cite{LPP+ComparingSMILESSELFIES2024,LFSMILESPairEncoding2021}.
While both tokenizers achieve full coverage of the OpenSMILES specification, differences between SMILES flavors can still lead to unknown tokens.
For example, MoleculeNet's HIV dataset includes \smiles{OCc1cc[te]c1}; however, \tok{[te]} is not valid per OpenSMILES\cite{CraOpenSMILES2016}.
As the only unknown token across our corpus of 2.1 billion molecules, the impact of \tok{[te]} on Smirk and Smirk-GPE is likely negligible.
Ongoing efforts to standardize the SMILES language could help clarify and eliminate these ambiguities\cite{ApoBalsaCompactLine2022,VinIUPACSMILESSpecification2019}.

We demonstrated n-grams as a proxy language model for tokenizer evaluation, positioning them as an intermediate step between existing intrinsic and extrinsic metrics.
We found n-gram performance metrics to be highly correlated with both pretraining and downstream performance metrics of transformer-based molecular foundation models (\cref{fig:ngram_prog}).
Our results suggest unknown tokens are detrimental to downstream performance, particularly for datasets on which the tokenizer has limited coverage.
As a single-point evaluation, our analysis may underestimate the performance that could be achieved with additional hyperparameter tuning.

Ultimately, molecular foundation models must deliver reliable and accurate predictions across the entire molecular design space.
Current, molecular language models\cite{RBC+LargescaleChemicalLanguage2022,SLG+MolecularTransformerModel2019,LFSMILESPairEncoding2021,BAVPMolGPTMolecularGeneration2022,IDHBChemformerPretrainedTransformer2022,ASC+ChemBERTa2ChemicalFoundation2022,SVLRPredictionChemicalReaction2021,SPV+MappingSpaceChemical2021,SSB+LargeEncoderDecoderFamily2024,LPP+ComparingSMILESSELFIES2024,TCPermutationInvariantGraphtoSequence2022} employ tokenizers that inadvertently obscure atom-level information, triggering a potentially significant loss of information (\cref{fig:ngram_metrics}).
This risk is not purely theoretical.
Bracketed atoms encode features essential to clinically relevant pharmaceuticals (Amoxicillin, 18F-Flurodeoxyglucose and Cisplatin), industrial applications (Tricalcium Sillicate and Diammonium Phosphate) and foundational discoveries (Vaska's Complex and Werner Complexes).
For example, Cisplatin (\cref{fig:example_tokenization}) is an effective chemotherapy drug, but its stereoisomer is not\cite{GhoCisplatinFirstMetal2019}.
Thus, omitting the chiral marker would erase medically relevant stereochemical information.

However, current benchmarks~\cite{WRF+MoleculeNetBenchmarkMolecular2018} lack sufficient diversity to evaluate molecular foundation models across the full breadth of chemical space.
tmQM~\cite{BSTmQMDatasetQuantum2020} partially addresses this gap with its expanded coverage of elements and greater diversity of stereochemical centers.
Nonetheless, benchmarks with comprehensive coverage of the periodic table, isotopes, charged species, and uncommon bond types such as quadruple bonds remain absent.
This scarcity of robust, diverse, and chemically challenging benchmarks limits the credibility and generalizability of evaluations for molecular foundation models.
We encourage the broader scientific foundation modeling community to rigorously assess the scope of their benchmarks, and the cheminformatics community to revisit and expand existing datasets.

We have so far neglected the core questions of whether a chemistry-specific tokenizer is necessary or if general-purpose tokenizers are sufficient.
Our results suggest that chemistry-specific tokenizers may result in more robust models (\cref{fig:ngram_metrics}), but not necessarily higher quality (\cref{fig:prognostic}).
Atom-wise tokenization improves model interpretability by allowing researchers to probe interatomic attention maps directly\cite{SPV+MappingSpaceChemical2021}.
Smirk expands on this by allowing researchers to directly manipulate the information rich content of bracketed atoms, while fully eliminating the risk of out-of-vocabulary tokens.

A foundation model for chemistry must encode the entire breadth of chemical space or risk obscuring critical features.
Mitigating this risk demands transitioning to tokenizers capable of encoding the entirety of chemical space.
Smirk -- and numerous other open-vocabulary tokenizers\cite{LOG+RoBERTaRobustlyOptimized2019,DJP+Llama3Herd2024,OAA+GPT4TechnicalReport2024,GMH+GemmaOpenModels2024} -- meet this threshold, are permissively licensed and well documented.
The impact on model performance appears negligible, with clear advantages to robustness and interpretability.

\section{Methods}
In total, we evaluated thirty-four tokenizers covering multiple domains, tokenization methods and molecular encodings.
Tokenizers were retrieved from various repositories per their author instructions and integrated with minimal cosmetic modification into our analysis pipeline.
A detailed manifest is provided in the supporting information.

\paragraph{Datasets}
We trained our molecular foundation and n-gram models using REALSpace, a dataset of more than 50B synthetically accessible make-on-demand molecule provided by Enamine\cite{EnaREALSpace2024}.
For downstream evaluations we used MoleculeNet\cite{WRF+MoleculeNetBenchmarkMolecular2018}, a collection of benchmark spanning quantum mechanics, physical chemistry, clinical domains, that has emerged as the de facto benchmark for molecular foundation models\cite{CGRChemBERTaLargeScaleSelfSupervised2020,RBC+LargescaleChemicalLanguage2022,SSB+LargeEncoderDecoderFamily2024,IDHBChemformerPretrainedTransformer2022,LFSMILESPairEncoding2021,LPP+ComparingSMILESSELFIES2024,FSA+NeuralScalingDeep2023}.
Additionally, we constructed a dataset of OpenSMILES\cite{CraOpenSMILES2016} molecular encodings from tmQM\cite{BSTmQMDatasetQuantum2020}, a quantum chemistry dataset of 108k transition metal complexes.
Additional details on each dataset and their curation can be found in our supporting information.

\paragraph{Smirk-GPE}
The Smirk-GPE tokenizer was trained on 262,035,601 molecules from the training split of our pretraining dataset derived from Enamine REAL Space \cite{EnaREALSpace2024}.
To reduce the number of unique ``words'' the training algorithm considers, we split SMILES strings on rings, brackets, non-bonds, and bracketed atoms using a regular expression.
We set a target vocabulary size of 50k tokens; however, training stopped at 2.3k tokens after exhausting all possible merges.
We did not explore the impact of vocabulary size\cite{GMFindingOptimalVocabulary2020} or corpus size\cite{GCE+UnpackingTokenizationEvaluating2024} on tokenization performance.
A second variant, \emph{Smirk-GPE, (NMB)} that excluded merges with brackets (\smiles{[} or \smiles{]}), was also trained and evaluated for this work, with similar results.

\subsection{Tokenizer Coverage}
We evaluated thirty-four tokenizers for their ability to tokenize chemical primitives and molecules from two benchmark datasets: MoleculeNet\cite{WRF+MoleculeNetBenchmarkMolecular2018} and tmQM\cite{BSTmQMDatasetQuantum2020}.
Results for eighteen representative tokenizers are displayed in \cref{fig:oov_tokens}, with comprehensive details for all tokenizers provided in our \datadrop[data release].
Initially, we enumerated the 126 single-atom OpenSMILES strings, encompassing the 118 elements and eight aromatic symbols.
This set was subsequently extended to include isotopes, chiral symbols (\smiles{@} and \smiles{@@}), oxidation states, and combinations of these attributes (e.g., charged, chiral isotopes).
Tokenizers were tasked with parsing representative OpenSMILES fragments for each case (e.g., \smiles{[18C@H3+2]}) and returning their token ids.
Coverage was quantified as the proportion of molecules processed without the unknown token id appearing in the results.
Oxidation states served as proxies for charged atom states, balancing chemical permissibility with the OpenSMILES constraint to accommodate charges between -15 and +15\cite{CraOpenSMILES2016}.
Fully enumerating the OpenSMILES\cite{CraOpenSMILES2016} specification is computationally intractable due to the 28 trillion permutation of bracketed atoms; see our supporting information for a complete figuring of this statistic.

\subsection{Training Protocol}\label{sec:training_protocol}
We pretrained 18 encoder-only transformer models spanning 11 tokenizers and three distinct molecular encoding schemes, leveraging a Masked Language Modeling (MLM) objective\cite{LOG+RoBERTaRobustlyOptimized2019}.
As the baseline, we employed HuggingFace's RoBERTa-PreLayerNorm architecture, featuring 8 layers, 8 attention heads, a hidden size of 512, an intermediate size of 2048, and a maximum sequence length of 2048.
Excluding embedding layers, the model consists of 25 million parameters.
Optimization was performed using the FusedLamb optimizer\cite{YLR+LargeBatchOptimization2020} with a learning rate of \(1.6 \times 10^{-4}\).

Pretraining was conducted using the Enamine REAL Space\cite{EnaREALSpace2024} dataset, partitioned into training (80\%), validation (10\%), and test (10\%) splits.
Models were trained for 30,000 steps with an effective batch size of 8192, distributed across two A100 GPUs, resulting in a total dataset size of 245 million molecules.
Validation cross-entropy loss was assessed every 12 optimizer steps on a sample of 98,304 molecules drawn from the validation split.
Canonical SMILES or SELFIES representations were generated on-the-fly using RDkit\cite{GreRDKitOpensourceCheminformatics2024} or the SELFIES Python packages\cite{LPN+RecentAdvancesSelfreferencing2023};
any transcoding errors were backfilled with molecules from the appropriate dataset split.

To compute predictions, an embedding vector was generated as the final hidden state of the first token for each tokenized molecule.
This embedding vector was passed through a two-layer neural network, referred to as the task network, to produce task-specific outputs.
Finetuning was conducted for 100,000 steps using the AdamW optimizer, with a maximum learning rate of \(1.6\times10^{-4}\), on a single A40 GPU.
For most models, except those training on the QM9 dataset, convergence was achieved well before reaching this step limit.
All models were trained with an effective batch size of 128, though for certain dataset and model combinations, this was split into micro-batches to manage memory constraints.
Similar to pretraining, molecules were transcoded on-the-fly, with failures removed.
Practically, this was only a concern for the tmQM dataset, where 54\% of the molecules could not be transcoded to SELFIES due to the presence of enhanced stereochemistry.
After finetuning, each model was evaluated on the test split using the checkpoint with the lowest validation loss.
Quality metrics for all models were computed using MoleculeNet's preferred metrics\cite{WRF+MoleculeNetBenchmarkMolecular2018} are included in the supplementary information.
Additionally, checkpoints for all pretrained and finetuned models are provided in our \datadrop[data release].

\subsection{N-Gram Models}
N-gram models and statistics were generated using a distributed codebase written in Julia\cite{BEKSJuliaFreshApproach2017} for this work, and the code is included in our \datadrop[data release].
Add-one smoothing was applied to handle zero-count tokens due to its simplicity and the absence of additional hyperparameters\cite{JMSpeechLanguageProcessing2024}.
Bidirectional n-gram probabilities were calculated as the joint distribution of the preceding and succeeding \(n-1\) tokens, capturing a total context of \(2n-2\).
Marginal distributions were carefully computed using exact integer arithmetic until the final stages of calculation to minimize floating-point rounding errors.
Additional numerical considerations are detailed in the supporting information.

\begin{acknowledgement}
The authors acknowledge support from Los Alamos National Laboratory.
This research was supported in part through computational resources and services provided by Advanced Research Computing at the University of Michigan, Ann Arbor.
This work used the Delta system at National Center for Supercomputing Applications through allocation CTS180061 from the Advanced Cyberinfrastructure Coordination Ecosystem: Services \& Support (ACCESS) program, which is supported by U.S. National Science Foundation grants \#2138259, \#2138286, \#2138307, \#2137603, and \#2138296.
Additionally, AW is grateful for the support of Meta's AR/AV Battery Research Fellowship.
AB is supported by a Catalyst grant from the Michigan Institute for Computational Discovery and Engineering at the University of Michigan.
\end{acknowledgement}

\bibliography{references}

\providecommand{\latin}[1]{#1}
\makeatletter
\providecommand{\doi}
  {\begingroup\let\do\@makeother\dospecials
  \catcode`\{=1 \catcode`\}=2 \doi@aux}
\providecommand{\doi@aux}[1]{\endgroup\texttt{#1}}
\makeatother
\providecommand*\mcitethebibliography{\thebibliography}
\csname @ifundefined\endcsname{endmcitethebibliography}
  {\let\endmcitethebibliography\endthebibliography}{}
\begin{mcitethebibliography}{75}
\providecommand*\natexlab[1]{#1}
\providecommand*\mciteSetBstSublistMode[1]{}
\providecommand*\mciteSetBstMaxWidthForm[2]{}
\providecommand*\mciteBstWouldAddEndPuncttrue
  {\def\EndOfBibitem{\unskip.}}
\providecommand*\mciteBstWouldAddEndPunctfalse
  {\let\EndOfBibitem\relax}
\providecommand*\mciteSetBstMidEndSepPunct[3]{}
\providecommand*\mciteSetBstSublistLabelBeginEnd[3]{}
\providecommand*\EndOfBibitem{}
\mciteSetBstSublistMode{f}
\mciteSetBstMaxWidthForm{subitem}{(\alph{mcitesubitemcount})}
\mciteSetBstSublistLabelBeginEnd
  {\mcitemaxwidthsubitemform\space}
  {\relax}
  {\relax}

\bibitem[Annevelink \latin{et~al.}(2022)Annevelink, Kurchin, Muckley, Kavalsky,
  Hegde, Sulzer, Zhu, Pu, Farina, Johnson, Gandhi, Dave, Lin, Edelman,
  Ramsundar, Saal, Rackauckas, Shah, Meredig, and
  Viswanathan]{AKM+AutoMatAutomatedMaterials2022}
Annevelink,~E. \latin{et~al.}  {{AutoMat}}: {{Automated}} Materials Discovery
  for Electrochemical Systems. \emph{MRS Bulletin} \textbf{2022}, \emph{47},
  1036--1044\relax
\mciteBstWouldAddEndPuncttrue
\mciteSetBstMidEndSepPunct{\mcitedefaultmidpunct}
{\mcitedefaultendpunct}{\mcitedefaultseppunct}\relax
\EndOfBibitem
\bibitem[{Pyzer-Knapp} \latin{et~al.}(2022){Pyzer-Knapp}, Pitera, Staar,
  Takeda, Laino, Sanders, Sexton, Smith, and
  Curioni]{PPS+AcceleratingMaterialsDiscovery2022}
{Pyzer-Knapp},~E.~O.; Pitera,~J.~W.; Staar,~P. W.~J.; Takeda,~S.; Laino,~T.;
  Sanders,~D.~P.; Sexton,~J.; Smith,~J.~R.; Curioni,~A. Accelerating Materials
  Discovery Using Artificial Intelligence, High Performance Computing and
  Robotics. \emph{npj Comput Mater} \textbf{2022}, \emph{8}, 84\relax
\mciteBstWouldAddEndPuncttrue
\mciteSetBstMidEndSepPunct{\mcitedefaultmidpunct}
{\mcitedefaultendpunct}{\mcitedefaultseppunct}\relax
\EndOfBibitem
\bibitem[Schapin \latin{et~al.}(2023)Schapin, Majewski, {Varela-Rial}, Arroniz,
  and Fabritiis]{SMV+MachineLearningSmall2023}
Schapin,~N.; Majewski,~M.; {Varela-Rial},~A.; Arroniz,~C.; Fabritiis,~G.~D.
  Machine Learning Small Molecule Properties in Drug Discovery.
  \emph{Artificial Intelligence Chemistry} \textbf{2023}, \emph{1},
  100020\relax
\mciteBstWouldAddEndPuncttrue
\mciteSetBstMidEndSepPunct{\mcitedefaultmidpunct}
{\mcitedefaultendpunct}{\mcitedefaultseppunct}\relax
\EndOfBibitem
\bibitem[Zeng \latin{et~al.}(2022)Zeng, Xiang, Yu, Wang, Li, Nussinov, and
  Cheng]{ZXY+AccuratePredictionMolecular2022}
Zeng,~X.; Xiang,~H.; Yu,~L.; Wang,~J.; Li,~K.; Nussinov,~R.; Cheng,~F. Accurate
  Prediction of Molecular Properties and Drug Targets Using a Self-Supervised
  Image Representation Learning Framework. \emph{Nat Mach Intell}
  \textbf{2022}, \emph{4}, 1004--1016\relax
\mciteBstWouldAddEndPuncttrue
\mciteSetBstMidEndSepPunct{\mcitedefaultmidpunct}
{\mcitedefaultendpunct}{\mcitedefaultseppunct}\relax
\EndOfBibitem
\bibitem[Mistry \latin{et~al.}(2021)Mistry, Franco, Cooper, Roberts, and
  Viswanathan]{MFC+HowMachineLearning2021}
Mistry,~A.; Franco,~A.~A.; Cooper,~S.~J.; Roberts,~S.~A.; Viswanathan,~V. How
  {{Machine Learning Will Revolutionize Electrochemical Sciences}}. \emph{ACS
  Energy Lett.} \textbf{2021}, \emph{6}, 1422--1431\relax
\mciteBstWouldAddEndPuncttrue
\mciteSetBstMidEndSepPunct{\mcitedefaultmidpunct}
{\mcitedefaultendpunct}{\mcitedefaultseppunct}\relax
\EndOfBibitem
\bibitem[Zhu \latin{et~al.}(2023)Zhu, Nguyen, Xia, Frost, Xie, Viswanathan, and
  Smith]{ZNX+ImprovedEnvironmentalChemistry2023}
Zhu,~S.; Nguyen,~B.~H.; Xia,~Y.; Frost,~K.; Xie,~S.; Viswanathan,~V.;
  Smith,~J.~A. Improved Environmental Chemistry Property Prediction of
  Molecules with Graph Machine Learning. \emph{Green Chem.} \textbf{2023},
  \emph{25}, 6612--6617\relax
\mciteBstWouldAddEndPuncttrue
\mciteSetBstMidEndSepPunct{\mcitedefaultmidpunct}
{\mcitedefaultendpunct}{\mcitedefaultseppunct}\relax
\EndOfBibitem
\bibitem[Batzner \latin{et~al.}(2022)Batzner, Musaelian, Sun, Geiger, Mailoa,
  Kornbluth, Molinari, Smidt, and Kozinsky]{BMS+E3equivariantGraphNeural2022}
Batzner,~S.; Musaelian,~A.; Sun,~L.; Geiger,~M.; Mailoa,~J.~P.; Kornbluth,~M.;
  Molinari,~N.; Smidt,~T.~E.; Kozinsky,~B. E(3)-Equivariant Graph Neural
  Networks for Data-Efficient and Accurate Interatomic Potentials. \emph{Nat
  Commun} \textbf{2022}, \emph{13}, 2453\relax
\mciteBstWouldAddEndPuncttrue
\mciteSetBstMidEndSepPunct{\mcitedefaultmidpunct}
{\mcitedefaultendpunct}{\mcitedefaultseppunct}\relax
\EndOfBibitem
\bibitem[Zhang \latin{et~al.}(2022)Zhang, Bier, and
  Viswanathan]{ZBVPredictingElectrolyteConductivity2022}
Zhang,~Y.; Bier,~I.; Viswanathan,~V. Predicting {{Electrolyte Conductivity
  Directly}} from {{Molecular-Level Interactions}}. \emph{ACS Energy Lett.}
  \textbf{2022}, \emph{7}, 4061--4070\relax
\mciteBstWouldAddEndPuncttrue
\mciteSetBstMidEndSepPunct{\mcitedefaultmidpunct}
{\mcitedefaultendpunct}{\mcitedefaultseppunct}\relax
\EndOfBibitem
\bibitem[Phuthi \latin{et~al.}(2024)Phuthi, Yao, Batzner, Musaelian, Guan,
  Kozinsky, Cubuk, and Viswanathan]{PYB+AccurateSurfaceFiniteTemperature2024}
Phuthi,~M.~K.; Yao,~A.~M.; Batzner,~S.; Musaelian,~A.; Guan,~P.; Kozinsky,~B.;
  Cubuk,~E.~D.; Viswanathan,~V. Accurate {{Surface}} and {{Finite-Temperature
  Bulk Properties}} of {{Lithium Metal}} at {{Large Scales Using Machine
  Learning Interaction Potentials}}. \emph{ACS Omega} \textbf{2024}, \emph{9},
  10904--10912\relax
\mciteBstWouldAddEndPuncttrue
\mciteSetBstMidEndSepPunct{\mcitedefaultmidpunct}
{\mcitedefaultendpunct}{\mcitedefaultseppunct}\relax
\EndOfBibitem
\bibitem[Vaswani \latin{et~al.}(2017)Vaswani, Shazeer, Parmar, Uszkoreit,
  Jones, Gomez, Kaiser, and Polosukhin]{VSP+AttentionAllYou2017}
Vaswani,~A.; Shazeer,~N.; Parmar,~N.; Uszkoreit,~J.; Jones,~L.; Gomez,~A.~N.;
  Kaiser,~L.; Polosukhin,~I. Attention {{Is All You Need}}. 2017\relax
\mciteBstWouldAddEndPuncttrue
\mciteSetBstMidEndSepPunct{\mcitedefaultmidpunct}
{\mcitedefaultendpunct}{\mcitedefaultseppunct}\relax
\EndOfBibitem
\bibitem[Liu \latin{et~al.}(2019)Liu, Ott, Goyal, Du, Joshi, Chen, Levy, Lewis,
  Zettlemoyer, and Stoyanov]{LOG+RoBERTaRobustlyOptimized2019}
Liu,~Y.; Ott,~M.; Goyal,~N.; Du,~J.; Joshi,~M.; Chen,~D.; Levy,~O.; Lewis,~M.;
  Zettlemoyer,~L.; Stoyanov,~V. {{RoBERTa}}: {{A Robustly Optimized BERT
  Pretraining Approach}}. 2019\relax
\mciteBstWouldAddEndPuncttrue
\mciteSetBstMidEndSepPunct{\mcitedefaultmidpunct}
{\mcitedefaultendpunct}{\mcitedefaultseppunct}\relax
\EndOfBibitem
\bibitem[Brown \latin{et~al.}(2020)Brown, Mann, Ryder, Subbiah, Kaplan,
  Dhariwal, Neelakantan, Shyam, Sastry, Askell, Agarwal, {Herbert-Voss},
  Krueger, Henighan, Child, Ramesh, Ziegler, Wu, Winter, Hesse, Chen, Sigler,
  Litwin, Gray, Chess, Clark, Berner, McCandlish, Radford, Sutskever, and
  Amodei]{BMR+LanguageModelsAre2020}
Brown,~T. \latin{et~al.}  Language {{Models}} Are {{Few-Shot Learners}}.
  Advances in {{Neural Information Processing Systems}}. 2020; pp
  1877--1901\relax
\mciteBstWouldAddEndPuncttrue
\mciteSetBstMidEndSepPunct{\mcitedefaultmidpunct}
{\mcitedefaultendpunct}{\mcitedefaultseppunct}\relax
\EndOfBibitem
\bibitem[Soares \latin{et~al.}(2024)Soares, Shirasuna, Brazil, Cerqueira,
  Zubarev, and Schmidt]{SSB+LargeEncoderDecoderFamily2024}
Soares,~E.; Shirasuna,~V.; Brazil,~E.~V.; Cerqueira,~R.; Zubarev,~D.;
  Schmidt,~K. A {{Large Encoder-Decoder Family}} of {{Foundation Models For
  Chemical Language}}. 2024\relax
\mciteBstWouldAddEndPuncttrue
\mciteSetBstMidEndSepPunct{\mcitedefaultmidpunct}
{\mcitedefaultendpunct}{\mcitedefaultseppunct}\relax
\EndOfBibitem
\bibitem[Y{\"u}ksel \latin{et~al.}(2023)Y{\"u}ksel, Ulusoy, {\"U}nl{\"u}, and
  Do{\u g}an]{YUUDSELFormerMolecularRepresentation2023}
Y{\"u}ksel,~A.; Ulusoy,~E.; {\"U}nl{\"u},~A.; Do{\u g}an,~T. {{SELFormer}}:
  Molecular Representation Learning via {{SELFIES}} Language Models.
  \emph{Mach. Learn.: Sci. Technol.} \textbf{2023}, \emph{4}, 025035\relax
\mciteBstWouldAddEndPuncttrue
\mciteSetBstMidEndSepPunct{\mcitedefaultmidpunct}
{\mcitedefaultendpunct}{\mcitedefaultseppunct}\relax
\EndOfBibitem
\bibitem[Xu \latin{et~al.}(2023)Xu, Wang, and
  Barati~Farimani]{XWBTransPolymerTransformerbasedLanguage2023}
Xu,~C.; Wang,~Y.; Barati~Farimani,~A. {{TransPolymer}}: A {{Transformer-based}}
  Language Model for Polymer Property Predictions. \emph{npj Comput Mater}
  \textbf{2023}, \emph{9}, 1--14\relax
\mciteBstWouldAddEndPuncttrue
\mciteSetBstMidEndSepPunct{\mcitedefaultmidpunct}
{\mcitedefaultendpunct}{\mcitedefaultseppunct}\relax
\EndOfBibitem
\bibitem[Ross \latin{et~al.}(2022)Ross, Belgodere, Chenthamarakshan, Padhi,
  Mroueh, and Das]{RBC+LargescaleChemicalLanguage2022}
Ross,~J.; Belgodere,~B.; Chenthamarakshan,~V.; Padhi,~I.; Mroueh,~Y.; Das,~P.
  Large-Scale Chemical Language Representations Capture Molecular Structure and
  Properties. \emph{Nat Mach Intell} \textbf{2022}, \emph{4}, 1256--1264\relax
\mciteBstWouldAddEndPuncttrue
\mciteSetBstMidEndSepPunct{\mcitedefaultmidpunct}
{\mcitedefaultendpunct}{\mcitedefaultseppunct}\relax
\EndOfBibitem
\bibitem[Chithrananda \latin{et~al.}(2020)Chithrananda, Grand, and
  Ramsundar]{CGRChemBERTaLargeScaleSelfSupervised2020}
Chithrananda,~S.; Grand,~G.; Ramsundar,~B. {{ChemBERTa}}: {{Large-Scale
  Self-Supervised Pretraining}} for {{Molecular Property Prediction}}.
  2020\relax
\mciteBstWouldAddEndPuncttrue
\mciteSetBstMidEndSepPunct{\mcitedefaultmidpunct}
{\mcitedefaultendpunct}{\mcitedefaultseppunct}\relax
\EndOfBibitem
\bibitem[Bagal \latin{et~al.}(2022)Bagal, Aggarwal, Vinod, and
  Priyakumar]{BAVPMolGPTMolecularGeneration2022}
Bagal,~V.; Aggarwal,~R.; Vinod,~P.~K.; Priyakumar,~U.~D. {{MolGPT}}:
  {{Molecular Generation Using}} a {{Transformer-Decoder Model}}. \emph{J.
  Chem. Inf. Model.} \textbf{2022}, \emph{62}, 2064--2076\relax
\mciteBstWouldAddEndPuncttrue
\mciteSetBstMidEndSepPunct{\mcitedefaultmidpunct}
{\mcitedefaultendpunct}{\mcitedefaultseppunct}\relax
\EndOfBibitem
\bibitem[Bilodeau \latin{et~al.}(2022)Bilodeau, Jin, Jaakkola, Barzilay, and
  Jensen]{BJJ+GenerativeModelsMolecular2022}
Bilodeau,~C.; Jin,~W.; Jaakkola,~T.; Barzilay,~R.; Jensen,~K.~F. Generative
  Models for Molecular Discovery: {{Recent}} Advances and Challenges.
  \emph{WIREs Computational Molecular Science} \textbf{2022}, \emph{12},
  e1608\relax
\mciteBstWouldAddEndPuncttrue
\mciteSetBstMidEndSepPunct{\mcitedefaultmidpunct}
{\mcitedefaultendpunct}{\mcitedefaultseppunct}\relax
\EndOfBibitem
\bibitem[Schwaller \latin{et~al.}(2019)Schwaller, Laino, Gaudin, Bolgar,
  Hunter, Bekas, and Lee]{SLG+MolecularTransformerModel2019}
Schwaller,~P.; Laino,~T.; Gaudin,~T.; Bolgar,~P.; Hunter,~C.~A.; Bekas,~C.;
  Lee,~A.~A. Molecular {{Transformer}}: {{A Model}} for
  {{Uncertainty-Calibrated Chemical Reaction Prediction}}. \emph{ACS Cent Sci}
  \textbf{2019}, \emph{5}, 1572--1583\relax
\mciteBstWouldAddEndPuncttrue
\mciteSetBstMidEndSepPunct{\mcitedefaultmidpunct}
{\mcitedefaultendpunct}{\mcitedefaultseppunct}\relax
\EndOfBibitem
\bibitem[Schwaller \latin{et~al.}(2018)Schwaller, Gaudin, L{\'a}nyi, Bekas, and
  Laino]{SGL+FoundTranslationPredicting2018}
Schwaller,~P.; Gaudin,~T.; L{\'a}nyi,~D.; Bekas,~C.; Laino,~T. ``{{Found}} in
  {{Translation}}'': Predicting Outcomes of Complex Organic Chemistry Reactions
  Using Neural Sequence-to-Sequence Models. \emph{Chem. Sci.} \textbf{2018},
  \emph{9}, 6091--6098\relax
\mciteBstWouldAddEndPuncttrue
\mciteSetBstMidEndSepPunct{\mcitedefaultmidpunct}
{\mcitedefaultendpunct}{\mcitedefaultseppunct}\relax
\EndOfBibitem
\bibitem[Schwaller \latin{et~al.}(2021)Schwaller, Vaucher, Laino, and
  Reymond]{SVLRPredictionChemicalReaction2021}
Schwaller,~P.; Vaucher,~A.~C.; Laino,~T.; Reymond,~J.-L. Prediction of Chemical
  Reaction Yields Using Deep Learning. \emph{Mach. Learn.: Sci. Technol.}
  \textbf{2021}, \emph{2}, 015016\relax
\mciteBstWouldAddEndPuncttrue
\mciteSetBstMidEndSepPunct{\mcitedefaultmidpunct}
{\mcitedefaultendpunct}{\mcitedefaultseppunct}\relax
\EndOfBibitem
\bibitem[Ramos \latin{et~al.}(2024)Ramos, Collison, and
  White]{RCWReviewLargeLanguage2024}
Ramos,~M.~C.; Collison,~C.~J.; White,~A.~D. A {{Review}} of {{Large Language
  Models}} and {{Autonomous Agents}} in {{Chemistry}}. 2024\relax
\mciteBstWouldAddEndPuncttrue
\mciteSetBstMidEndSepPunct{\mcitedefaultmidpunct}
{\mcitedefaultendpunct}{\mcitedefaultseppunct}\relax
\EndOfBibitem
\bibitem[Mielke \latin{et~al.}(2021)Mielke, Alyafeai, Salesky, Raffel, Dey,
  Gall{\'e}, Raja, Si, Lee, Sagot, and Tan]{MAS+WordsCharactersBrief2021}
Mielke,~S.~J.; Alyafeai,~Z.; Salesky,~E.; Raffel,~C.; Dey,~M.; Gall{\'e},~M.;
  Raja,~A.; Si,~C.; Lee,~W.~Y.; Sagot,~B.; Tan,~S. Between Words and
  Characters: {{A Brief History}} of {{Open-Vocabulary Modeling}} and
  {{Tokenization}} in {{NLP}}. 2021\relax
\mciteBstWouldAddEndPuncttrue
\mciteSetBstMidEndSepPunct{\mcitedefaultmidpunct}
{\mcitedefaultendpunct}{\mcitedefaultseppunct}\relax
\EndOfBibitem
\bibitem[Brants \latin{et~al.}(2007)Brants, Popat, Xu, Och, and
  Dean]{BPX+LargeLanguageModels2007}
Brants,~T.; Popat,~A.~C.; Xu,~P.; Och,~F.~J.; Dean,~J. Large {{Language
  Models}} in {{Machine Translation}}. Proceedings of the 2007 {{Joint
  Conference}} on {{Empirical Methods}} in {{Natural Language Processing}} and
  {{Computational Natural Language Learning}}. Prague, Czech Republic, 2007; pp
  858--867\relax
\mciteBstWouldAddEndPuncttrue
\mciteSetBstMidEndSepPunct{\mcitedefaultmidpunct}
{\mcitedefaultendpunct}{\mcitedefaultseppunct}\relax
\EndOfBibitem
\bibitem[Devlin \latin{et~al.}(2019)Devlin, Chang, Lee, and
  Toutanova]{DCLTBERTPretrainingDeep2019}
Devlin,~J.; Chang,~M.-W.; Lee,~K.; Toutanova,~K. {{BERT}}: {{Pre-training}} of
  {{Deep Bidirectional Transformers}} for {{Language Understanding}}.
  2019\relax
\mciteBstWouldAddEndPuncttrue
\mciteSetBstMidEndSepPunct{\mcitedefaultmidpunct}
{\mcitedefaultendpunct}{\mcitedefaultseppunct}\relax
\EndOfBibitem
\bibitem[Sennrich \latin{et~al.}(2016)Sennrich, Haddow, and
  Birch]{SHBNeuralMachineTranslation2016}
Sennrich,~R.; Haddow,~B.; Birch,~A. Neural {{Machine Translation}} of {{Rare
  Words}} with {{Subword Units}}. 2016\relax
\mciteBstWouldAddEndPuncttrue
\mciteSetBstMidEndSepPunct{\mcitedefaultmidpunct}
{\mcitedefaultendpunct}{\mcitedefaultseppunct}\relax
\EndOfBibitem
\bibitem[Hug(2024)]{HuggingfaceTokenizersFast2024}
Huggingface/Tokenizers: {{Fast State-of-the-Art Tokenizers}} Optimized for
  {{Research}} and {{Production}}. 2024\relax
\mciteBstWouldAddEndPuncttrue
\mciteSetBstMidEndSepPunct{\mcitedefaultmidpunct}
{\mcitedefaultendpunct}{\mcitedefaultseppunct}\relax
\EndOfBibitem
\bibitem[Land and Bartolo(2024)Land, and
  Bartolo]{LBFishingMagikarpAutomatically2024}
Land,~S.; Bartolo,~M. Fishing for {{Magikarp}}: {{Automatically Detecting
  Under-trained Tokens}} in {{Large Language Models}}. Proceedings of the 2024
  {{Conference}} on {{Empirical Methods}} in {{Natural Language Processing}}.
  Miami, Florida, USA, 2024; pp 11631--11646\relax
\mciteBstWouldAddEndPuncttrue
\mciteSetBstMidEndSepPunct{\mcitedefaultmidpunct}
{\mcitedefaultendpunct}{\mcitedefaultseppunct}\relax
\EndOfBibitem
\bibitem[Gage(1994)]{GagNewAlgorithmData1994}
Gage,~P. A New Algorithm for Data Compression. \emph{C Users J.} \textbf{1994},
  \emph{12}, 23--38\relax
\mciteBstWouldAddEndPuncttrue
\mciteSetBstMidEndSepPunct{\mcitedefaultmidpunct}
{\mcitedefaultendpunct}{\mcitedefaultseppunct}\relax
\EndOfBibitem
\bibitem[Schuster and Nakajima(2012)Schuster, and
  Nakajima]{SNJapaneseKoreanVoice2012}
Schuster,~M.; Nakajima,~K. Japanese and {{Korean}} Voice Search. {{ICASSP}}
  2012 - 2012 {{IEEE International Conference}} on {{Acoustics}}, {{Speech}}
  and {{Signal Processing}}. Kyoto, Japan, 2012; pp 5149--5152\relax
\mciteBstWouldAddEndPuncttrue
\mciteSetBstMidEndSepPunct{\mcitedefaultmidpunct}
{\mcitedefaultendpunct}{\mcitedefaultseppunct}\relax
\EndOfBibitem
\bibitem[Kudo(2018)]{KudSubwordRegularizationImproving2018}
Kudo,~T. Subword {{Regularization}}: {{Improving Neural Network Translation
  Models}} with {{Multiple Subword Candidates}}. Proceedings of the 56th
  {{Annual Meeting}} of the {{Association}} for {{Computational Linguistics}}
  ({{Volume}} 1: {{Long Papers}}). Melbourne, Australia, 2018; pp 66--75\relax
\mciteBstWouldAddEndPuncttrue
\mciteSetBstMidEndSepPunct{\mcitedefaultmidpunct}
{\mcitedefaultendpunct}{\mcitedefaultseppunct}\relax
\EndOfBibitem
\bibitem[Sagawa and Kojima(2023)Sagawa, and
  Kojima]{SKReactionT5LargescalePretrained2023}
Sagawa,~T.; Kojima,~R. {{ReactionT5}}: A Large-Scale Pre-Trained Model towards
  Application of Limited Reaction Data. 2023\relax
\mciteBstWouldAddEndPuncttrue
\mciteSetBstMidEndSepPunct{\mcitedefaultmidpunct}
{\mcitedefaultendpunct}{\mcitedefaultseppunct}\relax
\EndOfBibitem
\bibitem[Schwaller \latin{et~al.}(2021)Schwaller, Probst, Vaucher, Nair,
  Kreutter, Laino, and Reymond]{SPV+MappingSpaceChemical2021}
Schwaller,~P.; Probst,~D.; Vaucher,~A.~C.; Nair,~V.~H.; Kreutter,~D.;
  Laino,~T.; Reymond,~J.-L. Mapping the Space of Chemical Reactions Using
  Attention-Based Neural Networks. \emph{Nat Mach Intell} \textbf{2021},
  \emph{3}, 144--152\relax
\mciteBstWouldAddEndPuncttrue
\mciteSetBstMidEndSepPunct{\mcitedefaultmidpunct}
{\mcitedefaultendpunct}{\mcitedefaultseppunct}\relax
\EndOfBibitem
\bibitem[Irwin \latin{et~al.}(2022)Irwin, Dimitriadis, He, and
  Bjerrum]{IDHBChemformerPretrainedTransformer2022}
Irwin,~R.; Dimitriadis,~S.; He,~J.; Bjerrum,~E.~J. Chemformer: A Pre-Trained
  Transformer for Computational Chemistry. \emph{Mach. Learn.: Sci. Technol.}
  \textbf{2022}, \emph{3}, 015022\relax
\mciteBstWouldAddEndPuncttrue
\mciteSetBstMidEndSepPunct{\mcitedefaultmidpunct}
{\mcitedefaultendpunct}{\mcitedefaultseppunct}\relax
\EndOfBibitem
\bibitem[Bagal and Aggarwal(2023)Bagal, and Aggarwal]{BADevalabMolgpt2023}
Bagal,~V.; Aggarwal,~R. Devalab/Molgpt. 2023\relax
\mciteBstWouldAddEndPuncttrue
\mciteSetBstMidEndSepPunct{\mcitedefaultmidpunct}
{\mcitedefaultendpunct}{\mcitedefaultseppunct}\relax
\EndOfBibitem
\bibitem[Weininger(1988)]{WeiSMILESChemicalLanguage1988}
Weininger,~D. {{SMILES}}, a Chemical Language and Information System. 1.
  {{Introduction}} to Methodology and Encoding Rules. \emph{J. Chem. Inf.
  Comput. Sci.} \textbf{1988}, \emph{28}, 31--36\relax
\mciteBstWouldAddEndPuncttrue
\mciteSetBstMidEndSepPunct{\mcitedefaultmidpunct}
{\mcitedefaultendpunct}{\mcitedefaultseppunct}\relax
\EndOfBibitem
\bibitem[{Craig A. James}(2016)]{CraOpenSMILES2016}
{Craig A. James}, {{OpenSMILES}}. 2016\relax
\mciteBstWouldAddEndPuncttrue
\mciteSetBstMidEndSepPunct{\mcitedefaultmidpunct}
{\mcitedefaultendpunct}{\mcitedefaultseppunct}\relax
\EndOfBibitem
\bibitem[Krenn \latin{et~al.}(2020)Krenn, H{\"a}se, Nigam, Friederich, and
  {Aspuru-Guzik}]{KHN+SelfreferencingEmbeddedStrings2020}
Krenn,~M.; H{\"a}se,~F.; Nigam,~A.; Friederich,~P.; {Aspuru-Guzik},~A.
  Self-Referencing Embedded Strings ({{SELFIES}}): {{A}} 100\% Robust Molecular
  String Representation. \emph{Mach. Learn.: Sci. Technol.} \textbf{2020},
  \emph{1}, 045024\relax
\mciteBstWouldAddEndPuncttrue
\mciteSetBstMidEndSepPunct{\mcitedefaultmidpunct}
{\mcitedefaultendpunct}{\mcitedefaultseppunct}\relax
\EndOfBibitem
\bibitem[Li and Fourches(2021)Li, and Fourches]{LFSMILESPairEncoding2021}
Li,~X.; Fourches,~D. {{SMILES Pair Encoding}}: {{A Data-Driven Substructure
  Tokenization Algorithm}} for {{Deep Learning}}. \emph{J. Chem. Inf. Model.}
  \textbf{2021}, \emph{61}, 1560--1569\relax
\mciteBstWouldAddEndPuncttrue
\mciteSetBstMidEndSepPunct{\mcitedefaultmidpunct}
{\mcitedefaultendpunct}{\mcitedefaultseppunct}\relax
\EndOfBibitem
\bibitem[Leon \latin{et~al.}(2024)Leon, Perezhohin, Peres, Popovi{\v c}, and
  Castelli]{LPP+ComparingSMILESSELFIES2024}
Leon,~M.; Perezhohin,~Y.; Peres,~F.; Popovi{\v c},~A.; Castelli,~M. Comparing
  {{SMILES}} and {{SELFIES}} Tokenization for Enhanced Chemical Language
  Modeling. \emph{Sci Rep} \textbf{2024}, \emph{14}, 25016\relax
\mciteBstWouldAddEndPuncttrue
\mciteSetBstMidEndSepPunct{\mcitedefaultmidpunct}
{\mcitedefaultendpunct}{\mcitedefaultseppunct}\relax
\EndOfBibitem
\bibitem[Cherry \latin{et~al.}(2018)Cherry, Foster, Bapna, Firat, and
  Macherey]{CFB+RevisitingCharacterBasedNeural2018}
Cherry,~C.; Foster,~G.; Bapna,~A.; Firat,~O.; Macherey,~W. Revisiting
  {{Character-Based Neural Machine Translation}} with {{Capacity}} and
  {{Compression}}. Proceedings of the 2018 {{Conference}} on {{Empirical
  Methods}} in {{Natural Language Processing}}. Brussels, Belgium, 2018; pp
  4295--4305\relax
\mciteBstWouldAddEndPuncttrue
\mciteSetBstMidEndSepPunct{\mcitedefaultmidpunct}
{\mcitedefaultendpunct}{\mcitedefaultseppunct}\relax
\EndOfBibitem
\bibitem[{OpenAI}(2024)]{OpeOpenaiTiktoken2024}
{OpenAI}, Openai/Tiktoken. OpenAI, 2024\relax
\mciteBstWouldAddEndPuncttrue
\mciteSetBstMidEndSepPunct{\mcitedefaultmidpunct}
{\mcitedefaultendpunct}{\mcitedefaultseppunct}\relax
\EndOfBibitem
\bibitem[{National Center for Biotechnology Information}(2025)]{cid101490041}
{National Center for Biotechnology Information}, {{PubChem Compound Summary}}
  for {{CID}} 101490041, 1-{{Methyl-5-phenylsulfanylpyrazole}}. 2025\relax
\mciteBstWouldAddEndPuncttrue
\mciteSetBstMidEndSepPunct{\mcitedefaultmidpunct}
{\mcitedefaultendpunct}{\mcitedefaultseppunct}\relax
\EndOfBibitem
\bibitem[Batsuren \latin{et~al.}(2024)Batsuren, Vylomova, Dankers,
  Delgerbaatar, Uzan, Pinter, and Bella]{BVD+EvaluatingSubwordTokenization2024}
Batsuren,~K.; Vylomova,~E.; Dankers,~V.; Delgerbaatar,~T.; Uzan,~O.;
  Pinter,~Y.; Bella,~G. Evaluating {{Subword Tokenization}}: {{Alien Subword
  Composition}} and {{OOV Generalization Challenge}}. 2024\relax
\mciteBstWouldAddEndPuncttrue
\mciteSetBstMidEndSepPunct{\mcitedefaultmidpunct}
{\mcitedefaultendpunct}{\mcitedefaultseppunct}\relax
\EndOfBibitem
\bibitem[Goldman \latin{et~al.}(2024)Goldman, Caciularu, Eyal, Cao, Szpektor,
  and Tsarfaty]{GCE+UnpackingTokenizationEvaluating2024}
Goldman,~O.; Caciularu,~A.; Eyal,~M.; Cao,~K.; Szpektor,~I.; Tsarfaty,~R.
  Unpacking {{Tokenization}}: {{Evaluating Text Compression}} and Its
  {{Correlation}} with {{Model Performance}}. 2024\relax
\mciteBstWouldAddEndPuncttrue
\mciteSetBstMidEndSepPunct{\mcitedefaultmidpunct}
{\mcitedefaultendpunct}{\mcitedefaultseppunct}\relax
\EndOfBibitem
\bibitem[Ali \latin{et~al.}(2024)Ali, Fromm, Thellmann, Rutmann, L{\"u}bbering,
  Leveling, Klug, Ebert, Doll, Buschhoff, Jain, Weber, Jurkschat, Abdelwahab,
  John, Suarez, Ostendorff, Weinbach, Sifa, Kesselheim, and
  {Flores-Herr}]{AFT+TokenizerChoiceLLM2024}
Ali,~M. \latin{et~al.}  Tokenizer {{Choice For LLM Training}}: {{Negligible}}
  or {{Crucial}}? 2024\relax
\mciteBstWouldAddEndPuncttrue
\mciteSetBstMidEndSepPunct{\mcitedefaultmidpunct}
{\mcitedefaultendpunct}{\mcitedefaultseppunct}\relax
\EndOfBibitem
\bibitem[Rust \latin{et~al.}(2021)Rust, Pfeiffer, Vuli{\'c}, Ruder, and
  Gurevych]{RPV+HowGoodYour2021}
Rust,~P.; Pfeiffer,~J.; Vuli{\'c},~I.; Ruder,~S.; Gurevych,~I. How {{Good}} Is
  {{Your Tokenizer}}? {{On}} the {{Monolingual Performance}} of {{Language
  Models}}. Proceedings of the 59th {{Annual Meeting}} of the {{Association}}
  for {{Computational Linguistics}} and the 11th {{International Joint
  Conference}} on {{Natural Language Processing}} ({{Volume}} 1: {{Long
  Papers}}). Online, 2021; pp 3118--3135\relax
\mciteBstWouldAddEndPuncttrue
\mciteSetBstMidEndSepPunct{\mcitedefaultmidpunct}
{\mcitedefaultendpunct}{\mcitedefaultseppunct}\relax
\EndOfBibitem
\bibitem[Singh and Strouse(2024)Singh, and
  Strouse]{SSTokenizationCountsImpact2024}
Singh,~A.~K.; Strouse,~D.~J. Tokenization Counts: The Impact of Tokenization on
  Arithmetic in Frontier {{LLMs}}. 2024\relax
\mciteBstWouldAddEndPuncttrue
\mciteSetBstMidEndSepPunct{\mcitedefaultmidpunct}
{\mcitedefaultendpunct}{\mcitedefaultseppunct}\relax
\EndOfBibitem
\bibitem[Ahia \latin{et~al.}(2023)Ahia, Kumar, Gonen, Kasai, Mortensen, Smith,
  and Tsvetkov]{AKG+AllLanguagesCost2023}
Ahia,~O.; Kumar,~S.; Gonen,~H.; Kasai,~J.; Mortensen,~D.; Smith,~N.;
  Tsvetkov,~Y. Do {{All Languages Cost}} the {{Same}}? {{Tokenization}} in the
  {{Era}} of {{Commercial Language Models}}. Proceedings of the 2023
  {{Conference}} on {{Empirical Methods}} in {{Natural Language Processing}}.
  Singapore, 2023; pp 9904--9923\relax
\mciteBstWouldAddEndPuncttrue
\mciteSetBstMidEndSepPunct{\mcitedefaultmidpunct}
{\mcitedefaultendpunct}{\mcitedefaultseppunct}\relax
\EndOfBibitem
\bibitem[Lindsey \latin{et~al.}(2024)Lindsey, Pershing, Habib, Stephens,
  Blaschke, and Sundar]{LPH+ComparisonTokenizationImpact2024}
Lindsey,~L.~M.; Pershing,~N.~L.; Habib,~A.; Stephens,~W.~Z.; Blaschke,~A.~J.;
  Sundar,~H. A {{Comparison}} of {{Tokenization Impact}} in {{Attention Based}}
  and {{State Space Genomic Language Models}}. 2024\relax
\mciteBstWouldAddEndPuncttrue
\mciteSetBstMidEndSepPunct{\mcitedefaultmidpunct}
{\mcitedefaultendpunct}{\mcitedefaultseppunct}\relax
\EndOfBibitem
\bibitem[{Judit {\'A}cs}(2019)]{JudExploringBERTVocabulary2019}
{Judit {\'A}cs}, Exploring {{BERT}}'s {{Vocabulary}}. 2019\relax
\mciteBstWouldAddEndPuncttrue
\mciteSetBstMidEndSepPunct{\mcitedefaultmidpunct}
{\mcitedefaultendpunct}{\mcitedefaultseppunct}\relax
\EndOfBibitem
\bibitem[Gowda and May(2020)Gowda, and May]{GMFindingOptimalVocabulary2020}
Gowda,~T.; May,~J. Finding the {{Optimal Vocabulary Size}} for {{Neural Machine
  Translation}}. Findings of the {{Association}} for {{Computational
  Linguistics}}: {{EMNLP}} 2020. Online, 2020; pp 3955--3964\relax
\mciteBstWouldAddEndPuncttrue
\mciteSetBstMidEndSepPunct{\mcitedefaultmidpunct}
{\mcitedefaultendpunct}{\mcitedefaultseppunct}\relax
\EndOfBibitem
\bibitem[{Enamine Ltd.}(2024)]{EnaREALSpace2024}
{Enamine Ltd.}, {{REAL Space}}. 2024\relax
\mciteBstWouldAddEndPuncttrue
\mciteSetBstMidEndSepPunct{\mcitedefaultmidpunct}
{\mcitedefaultendpunct}{\mcitedefaultseppunct}\relax
\EndOfBibitem
\bibitem[Wu \latin{et~al.}(2018)Wu, Ramsundar, Feinberg, Gomes, Geniesse,
  Pappu, Leswing, and Pande]{WRF+MoleculeNetBenchmarkMolecular2018}
Wu,~Z.; Ramsundar,~B.; Feinberg,~E.~N.; Gomes,~J.; Geniesse,~C.; Pappu,~A.~S.;
  Leswing,~K.; Pande,~V. {{MoleculeNet}}: A Benchmark for Molecular Machine
  Learning. \emph{Chem. Sci.} \textbf{2018}, \emph{9}, 513--530\relax
\mciteBstWouldAddEndPuncttrue
\mciteSetBstMidEndSepPunct{\mcitedefaultmidpunct}
{\mcitedefaultendpunct}{\mcitedefaultseppunct}\relax
\EndOfBibitem
\bibitem[Balcells and Skjelstad(2020)Balcells, and
  Skjelstad]{BSTmQMDatasetQuantum2020}
Balcells,~D.; Skjelstad,~B.~B. {{tmQM Dataset}}---{{Quantum Geometries}} and
  {{Properties}} of 86k {{Transition Metal Complexes}}. \emph{J. Chem. Inf.
  Model.} \textbf{2020}, \emph{60}, 6135--6146\relax
\mciteBstWouldAddEndPuncttrue
\mciteSetBstMidEndSepPunct{\mcitedefaultmidpunct}
{\mcitedefaultendpunct}{\mcitedefaultseppunct}\relax
\EndOfBibitem
\bibitem[Shannon(1948)]{ShaMathematicalTheoryCommunication1948}
Shannon,~C.~E. A {{Mathematical Theory}} of {{Communication}}. \emph{The Bell
  System Technical Journal} \textbf{1948}, \emph{27}, 379--423\relax
\mciteBstWouldAddEndPuncttrue
\mciteSetBstMidEndSepPunct{\mcitedefaultmidpunct}
{\mcitedefaultendpunct}{\mcitedefaultseppunct}\relax
\EndOfBibitem
\bibitem[{Allen R. Wilcox}(1967)]{AllIndicesQualitativeVariation1967}
{Allen R. Wilcox}, \emph{Indices of {{Qualitative Variation}}}; 1967\relax
\mciteBstWouldAddEndPuncttrue
\mciteSetBstMidEndSepPunct{\mcitedefaultmidpunct}
{\mcitedefaultendpunct}{\mcitedefaultseppunct}\relax
\EndOfBibitem
\bibitem[Shannon(1993)]{ShaPredictionEntropyPrinted1993}
Shannon,~C.~E. In \emph{Claude {{E}}. {{Shannon}}: {{Collected Papers}}};
  Sloane,~N. J.~A., Wyner,~A.~D., Eds.; Wiley-IEEE Press, 1993; pp
  194--208\relax
\mciteBstWouldAddEndPuncttrue
\mciteSetBstMidEndSepPunct{\mcitedefaultmidpunct}
{\mcitedefaultendpunct}{\mcitedefaultseppunct}\relax
\EndOfBibitem
\bibitem[Workshop \latin{et~al.}(2023)Workshop, Scao, Fan, Akiki, Pavlick,
  Ili{\'c}, Hesslow, Castagn{\'e}, Luccioni, Yvon, Gall{\'e}, Tow, Rush,
  Biderman, Webson, Ammanamanchi, Wang, Sagot, Muennighoff, {del Moral},
  Ruwase, Bawden, Bekman, {McMillan-Major}, Beltagy, Nguyen, Saulnier, Tan,
  Suarez, Sanh, Lauren{\c c}on, Jernite, Launay, Mitchell, Raffel, Gokaslan,
  Simhi, Soroa, Aji, Alfassy, Rogers, Nitzav, Xu, Mou, Emezue, Klamm, Leong,
  {van Strien}, Adelani, Radev, Ponferrada, Levkovizh, Kim, Natan, De~Toni,
  Dupont, Kruszewski, Pistilli, Elsahar, Benyamina, Tran, Yu, Abdulmumin,
  Johnson, {Gonzalez-Dios}, {de la Rosa}, Chim, Dodge, Zhu, Chang, Frohberg,
  Tobing, Bhattacharjee, Almubarak, Chen, Lo, Von~Werra, Weber, Phan, {allal},
  Tanguy, Dey, Mu{\~n}oz, Masoud, Grandury, {\v S}a{\v s}ko, Huang, Coavoux,
  Singh, Jiang, Vu, Jauhar, Ghaleb, Subramani, Kassner, Khamis, Nguyen,
  Espejel, {de Gibert}, Villegas, Henderson, Colombo, Amuok, Lhoest, Harliman,
  Bommasani, L{\'o}pez, Ribeiro, Osei, Pyysalo, Nagel, Bose, Muhammad, Sharma,
  Longpre, Nikpoor, Silberberg, Pai, Zink, Torrent, Schick, Thrush, Danchev,
  Nikoulina, Laippala, Lepercq, Prabhu, Alyafeai, Talat, Raja, Heinzerling, Si,
  Ta{\c s}ar, Salesky, Mielke, Lee, Sharma, Santilli, Chaffin, Stiegler, Datta,
  Szczechla, Chhablani, Wang, Pandey, Strobelt, Fries, Rozen, Gao, Sutawika,
  Bari, {Al-shaibani}, Manica, Nayak, Teehan, Albanie, Shen, {Ben-David}, Bach,
  Kim, Bers, Fevry, Neeraj, Thakker, Raunak, Tang, Yong, Sun, Brody, Uri,
  Tojarieh, Roberts, Chung, Tae, Phang, Press, Li, Narayanan, Bourfoune,
  Casper, Rasley, Ryabinin, Mishra, Zhang, Shoeybi, Peyrounette, Patry, Tazi,
  Sanseviero, {von Platen}, Cornette, Lavall{\'e}e, Lacroix, Rajbhandari,
  Gandhi, Smith, Requena, Patil, Dettmers, Baruwa, Singh, Cheveleva, Ligozat,
  Subramonian, N{\'e}v{\'e}ol, Lovering, Garrette, Tunuguntla, Reiter,
  Taktasheva, Voloshina, Bogdanov, Winata, Schoelkopf, Kalo, Novikova, Forde,
  Clive, Kasai, Kawamura, Hazan, Carpuat, Clinciu, Kim, Cheng, Serikov,
  Antverg, {van der Wal}, Zhang, Zhang, Gehrmann, Mirkin, Pais, Shavrina,
  Scialom, Yun, Limisiewicz, Rieser, Protasov, Mikhailov, Pruksachatkun,
  Belinkov, Bamberger, Kasner, Rueda, Pestana, Feizpour, Khan, Faranak, Santos,
  Hevia, Unldreaj, Aghagol, Abdollahi, Tammour, HajiHosseini, Behroozi,
  Ajibade, Saxena, Ferrandis, McDuff, Contractor, Lansky, David, Kiela, Nguyen,
  Tan, Baylor, Ozoani, Mirza, Ononiwu, Rezanejad, Jones, Bhattacharya,
  Solaiman, Sedenko, Nejadgholi, Passmore, Seltzer, Sanz, Dutra, Samagaio,
  Elbadri, Mieskes, Gerchick, Akinlolu, McKenna, Qiu, Ghauri, Burynok, Abrar,
  Rajani, Elkott, Fahmy, Samuel, An, Kromann, Hao, Alizadeh, Shubber, Wang,
  Roy, Viguier, Le, Oyebade, Le, Yang, Nguyen, Kashyap, Palasciano, Callahan,
  Shukla, {Miranda-Escalada}, Singh, Beilharz, Wang, Brito, Zhou, Jain, Xu,
  Fourrier, Peri{\~n}{\'a}n, Molano, Yu, Manjavacas, Barth, Fuhrimann, Altay,
  Bayrak, Burns, Vrabec, Bello, Dash, Kang, Giorgi, Golde, Posada, Sivaraman,
  Bulchandani, Liu, Shinzato, {de Bykhovetz}, Takeuchi, P{\`a}mies, Castillo,
  Nezhurina, S{\"a}nger, Samwald, Cullan, Weinberg, De~Wolf, Mihaljcic, Liu,
  Freidank, Kang, Seelam, Dahlberg, Broad, Muellner, Fung, Haller,
  Chandrasekhar, Eisenberg, Martin, Canalli, Su, Su, Cahyawijaya, Garda,
  Deshmukh, Mishra, Kiblawi, Ott, {Sang-aroonsiri}, Kumar, Schweter, Bharati,
  Laud, Gigant, Kainuma, Kusa, Labrak, Bajaj, Venkatraman, Xu, Xu, Xu, Tan,
  Xie, Ye, Bras, Belkada, and Wolf]{WSF+BLOOM176BParameterOpenAccess2023}
Workshop,~B. \latin{et~al.}  {{BLOOM}}: {{A 176B-Parameter Open-Access
  Multilingual Language Model}}. 2023\relax
\mciteBstWouldAddEndPuncttrue
\mciteSetBstMidEndSepPunct{\mcitedefaultmidpunct}
{\mcitedefaultendpunct}{\mcitedefaultseppunct}\relax
\EndOfBibitem
\bibitem[Apodaca(2022)]{ApoBalsaCompactLine2022}
Apodaca,~R. Balsa: {{A Compact Line Notation Based}} on {{SMILES}}. 2022\relax
\mciteBstWouldAddEndPuncttrue
\mciteSetBstMidEndSepPunct{\mcitedefaultmidpunct}
{\mcitedefaultendpunct}{\mcitedefaultseppunct}\relax
\EndOfBibitem
\bibitem[{Vincent F Scalfani}(2019)]{VinIUPACSMILESSpecification2019}
{Vincent F Scalfani}, {{IUPAC SMILES}}+ {{Specification}}. 2019\relax
\mciteBstWouldAddEndPuncttrue
\mciteSetBstMidEndSepPunct{\mcitedefaultmidpunct}
{\mcitedefaultendpunct}{\mcitedefaultseppunct}\relax
\EndOfBibitem
\bibitem[Ahmad \latin{et~al.}(2022)Ahmad, Simon, Chithrananda, Grand, and
  Ramsundar]{ASC+ChemBERTa2ChemicalFoundation2022}
Ahmad,~W.; Simon,~E.; Chithrananda,~S.; Grand,~G.; Ramsundar,~B.
  {{ChemBERTa-2}}: {{Towards Chemical Foundation Models}}. 2022\relax
\mciteBstWouldAddEndPuncttrue
\mciteSetBstMidEndSepPunct{\mcitedefaultmidpunct}
{\mcitedefaultendpunct}{\mcitedefaultseppunct}\relax
\EndOfBibitem
\bibitem[Tu and Coley(2022)Tu, and
  Coley]{TCPermutationInvariantGraphtoSequence2022}
Tu,~Z.; Coley,~C.~W. Permutation {{Invariant Graph-to-Sequence Model}} for
  {{Template-Free Retrosynthesis}} and {{Reaction Prediction}}. \emph{J. Chem.
  Inf. Model.} \textbf{2022}, \emph{62}, 3503--3513\relax
\mciteBstWouldAddEndPuncttrue
\mciteSetBstMidEndSepPunct{\mcitedefaultmidpunct}
{\mcitedefaultendpunct}{\mcitedefaultseppunct}\relax
\EndOfBibitem
\bibitem[Ghosh(2019)]{GhoCisplatinFirstMetal2019}
Ghosh,~S. Cisplatin: {{The}} First Metal Based Anticancer Drug.
  \emph{Bioorganic Chemistry} \textbf{2019}, \emph{88}, 102925\relax
\mciteBstWouldAddEndPuncttrue
\mciteSetBstMidEndSepPunct{\mcitedefaultmidpunct}
{\mcitedefaultendpunct}{\mcitedefaultseppunct}\relax
\EndOfBibitem
\bibitem[Dubey \latin{et~al.}(2024)Dubey, Jauhri, Pandey, Kadian, {Al-Dahle},
  Letman, Mathur, Schelten, Yang, Fan, Goyal, Hartshorn, Yang, Mitra,
  Sravankumar, Korenev, Hinsvark, Rao, Zhang, Rodriguez, Gregerson, Spataru,
  Roziere, Biron, Tang, Chern, Caucheteux, Nayak, Bi, Marra, McConnell, Keller,
  Touret, Wu, Wong, Ferrer, Nikolaidis, Allonsius, Song, Pintz, Livshits,
  Esiobu, Choudhary, Mahajan, {Garcia-Olano}, Perino, Hupkes, Lakomkin,
  AlBadawy, Lobanova, Dinan, Smith, Radenovic, Zhang, Synnaeve, Lee, Anderson,
  Nail, Mialon, Pang, Cucurell, Nguyen, Korevaar, Xu, Touvron, Zarov, Ibarra,
  Kloumann, Misra, Evtimov, Copet, Lee, Geffert, Vranes, Park, Mahadeokar,
  Shah, {van der Linde}, Billock, Hong, Lee, Fu, Chi, Huang, Liu, Wang, Yu,
  Bitton, Spisak, Park, Rocca, Johnstun, Saxe, Jia, Alwala, Upasani, Plawiak,
  Li, Heafield, Stone, {El-Arini}, Iyer, Malik, Chiu, Bhalla, {Rantala-Yeary},
  {van der Maaten}, Chen, Tan, Jenkins, Martin, Madaan, Malo, Blecher,
  Landzaat, {de Oliveira}, Muzzi, Pasupuleti, Singh, Paluri, Kardas, Oldham,
  Rita, Pavlova, Kambadur, Lewis, Si, Singh, Hassan, Goyal, Torabi, Bashlykov,
  Bogoychev, Chatterji, Duchenne, {\c C}elebi, Alrassy, Zhang, Li, Vasic, Weng,
  Bhargava, Dubal, Krishnan, Koura, Xu, He, Dong, Srinivasan, Ganapathy,
  Calderer, Cabral, Stojnic, Raileanu, Girdhar, Patel, Sauvestre, Polidoro,
  Sumbaly, Taylor, Silva, Hou, Wang, Hosseini, Chennabasappa, Singh, Bell, Kim,
  Edunov, Nie, Narang, Raparthy, Shen, Wan, Bhosale, Zhang, Vandenhende, Batra,
  Whitman, Sootla, Collot, Gururangan, Borodinsky, Herman, Fowler, Sheasha,
  Georgiou, Scialom, Speckbacher, Mihaylov, Xiao, Karn, Goswami, Gupta,
  Ramanathan, Kerkez, Gonguet, Do, Vogeti, Petrovic, Chu, Xiong, Fu, Meers,
  Martinet, Wang, Tan, Xie, Jia, Wang, Goldschlag, Gaur, Babaei, Wen, Song,
  Zhang, Li, Mao, Coudert, Yan, Chen, Papakipos, Singh, Grattafiori, Jain,
  Kelsey, Shajnfeld, Gangidi, Victoria, Goldstand, Menon, Sharma, Boesenberg,
  Vaughan, Baevski, Feinstein, Kallet, Sangani, Yunus, Lupu, Alvarado, Caples,
  Gu, Ho, Poulton, Ryan, Ramchandani, Franco, Saraf, Chowdhury, Gabriel,
  Bharambe, Eisenman, Yazdan, James, Maurer, Leonhardi, Huang, Loyd, De~Paola,
  Paranjape, Liu, Wu, Ni, Hancock, Wasti, Spence, Stojkovic, Gamido, Montalvo,
  Parker, Burton, Mejia, Wang, Kim, Zhou, Hu, Chu, Cai, Tindal, Feichtenhofer,
  Civin, Beaty, Kreymer, Li, Wyatt, Adkins, Xu, Testuggine, David, Parikh,
  Liskovich, Foss, Wang, Le, Holland, Dowling, Jamil, Montgomery, Presani,
  Hahn, Wood, Brinkman, Arcaute, Dunbar, Smothers, Sun, Kreuk, Tian, Ozgenel,
  Caggioni, Guzm{\'a}n, Kanayet, Seide, Florez, Schwarz, Badeer, Swee, Halpern,
  Thattai, Herman, Sizov, Guangyi, Zhang, Lakshminarayanan, Shojanazeri, Zou,
  Wang, Zha, Habeeb, Rudolph, Suk, Aspegren, Goldman, Damlaj, Molybog, Tufanov,
  Veliche, Gat, Weissman, Geboski, Kohli, Asher, Gaya, Marcus, Tang, Chan,
  Zhen, Reizenstein, Teboul, Zhong, Jin, Yang, Cummings, Carvill, Shepard,
  McPhie, Torres, Ginsburg, Wang, Wu, U, Saxena, Prasad, Khandelwal, Zand,
  Matosich, Veeraraghavan, Michelena, Li, Huang, Chawla, Lakhotia, Huang, Chen,
  Garg, A, Silva, Bell, Zhang, Guo, Yu, Moshkovich, Wehrstedt, Khabsa, Avalani,
  Bhatt, Tsimpoukelli, Mankus, Hasson, Lennie, Reso, Groshev, Naumov, Lathi,
  Keneally, Seltzer, Valko, Restrepo, Patel, Vyatskov, Samvelyan, Clark, Macey,
  Wang, Hermoso, Metanat, Rastegari, Bansal, Santhanam, Parks, White, Bawa,
  Singhal, Egebo, Usunier, Laptev, Dong, Zhang, Cheng, Chernoguz, Hart,
  Salpekar, Kalinli, Kent, Parekh, Saab, Balaji, Rittner, Bontrager, Roux,
  Dollar, Zvyagina, Ratanchandani, Yuvraj, Liang, Alao, Rodriguez, Ayub,
  Murthy, Nayani, Mitra, Li, Hogan, Battey, Wang, Maheswari, Howes, Rinott,
  Bondu, Datta, Chugh, Hunt, Dhillon, Sidorov, Pan, Verma, Yamamoto, Ramaswamy,
  Lindsay, Lindsay, Feng, Lin, Zha, Shankar, Zhang, Zhang, Wang, Agarwal,
  Sajuyigbe, Chintala, Max, Chen, Kehoe, Satterfield, Govindaprasad, Gupta,
  Cho, Virk, Subramanian, Choudhury, Goldman, Remez, Glaser, Best, Kohler,
  Robinson, Li, Zhang, Matthews, Chou, Shaked, Vontimitta, Ajayi, Montanez,
  Mohan, Kumar, Mangla, Albiero, Ionescu, Poenaru, Mihailescu, Ivanov, Li,
  Wang, Jiang, Bouaziz, Constable, Tang, Wang, Wu, Wang, Xia, Wu, Gao, Chen,
  Hu, Jia, Qi, Li, Zhang, Zhang, Adi, Nam, Yu, Wang, Hao, Qian, He, Rait,
  DeVito, Rosnbrick, Wen, Yang, and Zhao]{DJP+Llama3Herd2024}
Dubey,~A. \latin{et~al.}  The {{Llama}} 3 {{Herd}} of {{Models}}. 2024\relax
\mciteBstWouldAddEndPuncttrue
\mciteSetBstMidEndSepPunct{\mcitedefaultmidpunct}
{\mcitedefaultendpunct}{\mcitedefaultseppunct}\relax
\EndOfBibitem
\bibitem[OpenAI \latin{et~al.}(2024)OpenAI, Achiam, Adler, Agarwal, Ahmad,
  Akkaya, Aleman, Almeida, Altenschmidt, Altman, Anadkat, Avila, Babuschkin,
  Balaji, Balcom, Baltescu, Bao, Bavarian, Belgum, Bello, Berdine,
  {Bernadett-Shapiro}, Berner, Bogdonoff, Boiko, Boyd, Brakman, Brockman,
  Brooks, Brundage, Button, Cai, Campbell, Cann, Carey, Carlson, Carmichael,
  Chan, Chang, Chantzis, Chen, Chen, Chen, Chen, Chen, Chess, Cho, Chu, Chung,
  Cummings, Currier, Dai, Decareaux, Degry, Deutsch, Deville, Dhar, Dohan,
  Dowling, Dunning, Ecoffet, Eleti, Eloundou, Farhi, Fedus, Felix, Fishman,
  Forte, Fulford, Gao, Georges, Gibson, Goel, Gogineni, Goh, {Gontijo-Lopes},
  Gordon, Grafstein, Gray, Greene, Gross, Gu, Guo, Hallacy, Han, Harris, He,
  Heaton, Heidecke, Hesse, Hickey, Hickey, Hoeschele, Houghton, Hsu, Hu, Hu,
  Huizinga, Jain, Jain, Jang, Jiang, Jiang, Jin, Jin, Jomoto, Jonn, Jun,
  Kaftan, Kaiser, Kamali, Kanitscheider, Keskar, Khan, Kilpatrick, Kim, Kim,
  Kim, Kirchner, Kiros, Knight, Kokotajlo, Kondraciuk, Kondrich,
  Konstantinidis, Kosic, Krueger, Kuo, Lampe, Lan, Lee, Leike, Leung, Levy, Li,
  Lim, Lin, Lin, Litwin, Lopez, Lowe, Lue, Makanju, Malfacini, Manning, Markov,
  Markovski, Martin, Mayer, Mayne, McGrew, McKinney, McLeavey, McMillan,
  McNeil, Medina, Mehta, Menick, Metz, Mishchenko, Mishkin, Monaco, Morikawa,
  Mossing, Mu, Murati, Murk, M{\'e}ly, Nair, Nakano, Nayak, Neelakantan, Ngo,
  Noh, Ouyang, O'Keefe, Pachocki, Paino, Palermo, Pantuliano, Parascandolo,
  Parish, Parparita, Passos, Pavlov, Peng, Perelman, Peres, Petrov, Pinto,
  Michael, Pokorny, Pokrass, Pong, Powell, Power, Power, Proehl, Puri, Radford,
  Rae, Ramesh, Raymond, Real, Rimbach, Ross, Rotsted, Roussez, Ryder,
  Saltarelli, Sanders, Santurkar, Sastry, Schmidt, Schnurr, Schulman, Selsam,
  Sheppard, Sherbakov, Shieh, Shoker, Shyam, Sidor, Sigler, Simens, Sitkin,
  Slama, Sohl, Sokolowsky, Song, Staudacher, Such, Summers, Sutskever, Tang,
  Tezak, Thompson, Tillet, Tootoonchian, Tseng, Tuggle, Turley, Tworek, Uribe,
  Vallone, Vijayvergiya, Voss, Wainwright, Wang, Wang, Wang, Ward, Wei,
  Weinmann, Welihinda, Welinder, Weng, Weng, Wiethoff, Willner, Winter,
  Wolrich, Wong, Workman, Wu, Wu, Wu, Xiao, Xu, Yoo, Yu, Yuan, Zaremba,
  Zellers, Zhang, Zhang, Zhao, Zheng, Zhuang, Zhuk, and
  Zoph]{OAA+GPT4TechnicalReport2024}
OpenAI, \latin{et~al.}  {{GPT-4 Technical Report}}. 2024\relax
\mciteBstWouldAddEndPuncttrue
\mciteSetBstMidEndSepPunct{\mcitedefaultmidpunct}
{\mcitedefaultendpunct}{\mcitedefaultseppunct}\relax
\EndOfBibitem
\bibitem[Team \latin{et~al.}(2024)Team, Mesnard, Hardin, Dadashi, Bhupatiraju,
  Pathak, Sifre, Rivi{\`e}re, Kale, Love, Tafti, Hussenot, Sessa, Chowdhery,
  Roberts, Barua, Botev, {Castro-Ros}, Slone, H{\'e}liou, Tacchetti, Bulanova,
  Paterson, Tsai, Shahriari, Lan, {Choquette-Choo}, Crepy, Cer, Ippolito, Reid,
  Buchatskaya, Ni, Noland, Yan, Tucker, Muraru, Rozhdestvenskiy, Michalewski,
  Tenney, Grishchenko, Austin, Keeling, Labanowski, Lespiau, Stanway, Brennan,
  Chen, Ferret, Chiu, {Mao-Jones}, Lee, Yu, Millican, Sjoesund, Lee, Dixon,
  Reid, Miku{\l}a, Wirth, Sharman, Chinaev, Thain, Bachem, Chang, Wahltinez,
  Bailey, Michel, Yotov, Chaabouni, Comanescu, Jana, Anil, McIlroy, Liu,
  Mullins, Smith, Borgeaud, Girgin, Douglas, Pandya, Shakeri, De, Klimenko,
  Hennigan, Feinberg, Stokowiec, Chen, Ahmed, Gong, Warkentin, Peran, Giang,
  Farabet, Vinyals, Dean, Kavukcuoglu, Hassabis, Ghahramani, Eck, Barral,
  Pereira, Collins, Joulin, Fiedel, Senter, Andreev, and
  Kenealy]{GMH+GemmaOpenModels2024}
Team,~G. \latin{et~al.}  Gemma: {{Open Models Based}} on {{Gemini Research}}
  and {{Technology}}. 2024\relax
\mciteBstWouldAddEndPuncttrue
\mciteSetBstMidEndSepPunct{\mcitedefaultmidpunct}
{\mcitedefaultendpunct}{\mcitedefaultseppunct}\relax
\EndOfBibitem
\bibitem[Frey \latin{et~al.}(2023)Frey, Soklaski, Axelrod, Samsi,
  {G{\'o}mez-Bombarelli}, Coley, and Gadepally]{FSA+NeuralScalingDeep2023}
Frey,~N.~C.; Soklaski,~R.; Axelrod,~S.; Samsi,~S.; {G{\'o}mez-Bombarelli},~R.;
  Coley,~C.~W.; Gadepally,~V. Neural Scaling of Deep Chemical Models. \emph{Nat
  Mach Intell} \textbf{2023}, \emph{5}, 1297--1305\relax
\mciteBstWouldAddEndPuncttrue
\mciteSetBstMidEndSepPunct{\mcitedefaultmidpunct}
{\mcitedefaultendpunct}{\mcitedefaultseppunct}\relax
\EndOfBibitem
\bibitem[You \latin{et~al.}(2020)You, Li, Reddi, Hseu, Kumar, Bhojanapalli,
  Song, Demmel, Keutzer, and Hsieh]{YLR+LargeBatchOptimization2020}
You,~Y.; Li,~J.; Reddi,~S.; Hseu,~J.; Kumar,~S.; Bhojanapalli,~S.; Song,~X.;
  Demmel,~J.; Keutzer,~K.; Hsieh,~C.-J. Large {{Batch Optimization}} for {{Deep
  Learning}}: {{Training BERT}} in 76 Minutes. 2020\relax
\mciteBstWouldAddEndPuncttrue
\mciteSetBstMidEndSepPunct{\mcitedefaultmidpunct}
{\mcitedefaultendpunct}{\mcitedefaultseppunct}\relax
\EndOfBibitem
\bibitem[{Greg Landrum}(2024)]{GreRDKitOpensourceCheminformatics2024}
{Greg Landrum}, {{RDKit}}: {{Open-source}} Cheminformatics. 2024\relax
\mciteBstWouldAddEndPuncttrue
\mciteSetBstMidEndSepPunct{\mcitedefaultmidpunct}
{\mcitedefaultendpunct}{\mcitedefaultseppunct}\relax
\EndOfBibitem
\bibitem[Lo \latin{et~al.}(2023)Lo, Pollice, Nigam, White, Krenn, and
  {Aspuru-Guzik}]{LPN+RecentAdvancesSelfreferencing2023}
Lo,~A.; Pollice,~R.; Nigam,~A.; White,~A.~D.; Krenn,~M.; {Aspuru-Guzik},~A.
  Recent Advances in the Self-Referencing Embedded Strings ({{SELFIES}})
  Library. \emph{Digital Discovery} \textbf{2023}, \emph{2}, 897--908\relax
\mciteBstWouldAddEndPuncttrue
\mciteSetBstMidEndSepPunct{\mcitedefaultmidpunct}
{\mcitedefaultendpunct}{\mcitedefaultseppunct}\relax
\EndOfBibitem
\bibitem[Bezanson \latin{et~al.}(2017)Bezanson, Edelman, Karpinski, and
  Shah]{BEKSJuliaFreshApproach2017}
Bezanson,~J.; Edelman,~A.; Karpinski,~S.; Shah,~V.~B. Julia: {{A}} Fresh
  Approach to Numerical Computing. \emph{SIAM Review} \textbf{2017}, \emph{59},
  65--98\relax
\mciteBstWouldAddEndPuncttrue
\mciteSetBstMidEndSepPunct{\mcitedefaultmidpunct}
{\mcitedefaultendpunct}{\mcitedefaultseppunct}\relax
\EndOfBibitem
\bibitem[Jurafsky and Martin(2024)Jurafsky, and
  Martin]{JMSpeechLanguageProcessing2024}
Jurafsky,~D.; Martin,~J.~H. \emph{Speech and Language Processing: {{An}}
  Introduction to Natural Language Processing, Computational Linguistics, and
  Speech Recognition with Language Models}, 3rd ed.; 2024\relax
\mciteBstWouldAddEndPuncttrue
\mciteSetBstMidEndSepPunct{\mcitedefaultmidpunct}
{\mcitedefaultendpunct}{\mcitedefaultseppunct}\relax
\EndOfBibitem
\end{mcitethebibliography}

\end{document}